\documentclass[conference]{IEEEtran}
\IEEEoverridecommandlockouts

\usepackage[english]{babel} 
\usepackage[T1]{fontenc} 
\usepackage[utf8]{inputenc} 
\usepackage{graphicx}

\usepackage[hyphens]{url}
\usepackage[nospace,noadjust]{cite}
\usepackage{amsmath,amsfonts,amsthm,amssymb}
\usepackage[caption=false,font=footnotesize,hangindent=10pt]{subfig}
\usepackage[linesnumbered,vlined,boxed,ruled]{algorithm2e}
\usepackage{color,xparse}
\usepackage{tabu}
\usepackage[colorlinks]{hyperref} 
\usepackage[capitalise]{cleveref}

\usepackage{makecell}

\usepackage{tikz}
\usetikzlibrary{shapes.geometric,shapes.misc,positioning,backgrounds,fit,arrows.meta}

\graphicspath{{./}{figure/}}
\def\Snospace~{\S{}}

\SetVlineSkip{2pt}

\SetCommentSty{myAlgComment}
\SetKwProg{Fn}{Function}{:}{}
\SetAlFnt{\small}
\SetAlCapFnt{\small}
\SetAlCapNameFnt{\small}

\newcommand{\header}[1]{\smallskip\noindent\textbf{#1}}

\NewDocumentCommand{\E}{o m}{\mathbb{E}\IfValueTF{#1}{_{#1}}{}\big(#2\big)}

\newcommand{\indr}[1]{\mathbf{1}(#1)}
\newcommand{\softmax}[1]{\mathrm{Softmax}(#1)}
\newcommand{\pr}[1]{\langle {#1} \rangle}
\newcommand{\Real}{\mathbb{R}}
\newcommand{\CA}{\mathcal{A}}
\newcommand{\CK}{\mathcal{K}}
\newcommand{\CV}{\mathcal{V}}
\newcommand{\CS}{\mathcal{S}}

\newcommand{\bv}{\mathbf{v}}
\newcommand{\bE}{\mathbf{E}}
\newcommand{\bQ}{\mathbf{Q}}
\newcommand{\bK}{\mathbf{K}}
\newcommand{\bV}{\mathbf{V}}
\newcommand{\bM}{\mathbf{M}}

\newcommand{\bW}{\mathbf{W}}
\newcommand{\bw}{\mathbf{w}}
\newcommand{\bb}{\mathbf{b}}

\newcommand{\bs}{\mathbf{s}}
\newcommand{\bp}{\mathbf{p}}
\newcommand{\btheta}{\boldsymbol{\theta}}

\title{Representation Learning of Tangled Key-Value Sequence Data for Early Classification}

\author{%
  \IEEEauthorblockN{Tao Duan\quad Junzhou Zhao\thanks{* Corresponding Author}\IEEEauthorrefmark{1}\quad Shuo Zhang\quad Jing Tao\quad Pinghui Wang}
  \IEEEauthorblockA{MOE KLINNS Lab, Xi'an Jiaotong University, Xi'an 710049, P.~R.~China\\
    \{duantao, zs412082986\}@stu.xjtu.edu.cn, \{junzhou.zhao, jtao, phwang\}@xjtu.edu.cn}%
}

\begin{document}

\maketitle

\begin{abstract}
Key-value sequence data has become ubiquitous and naturally appears in a variety
of real-world applications, ranging from the user-product purchasing sequences in
e-commerce, to network packet sequences forwarded by routers in networking.
Classifying these key-value sequences is important in many scenarios such as user
profiling and malicious applications identification.
In many time-sensitive scenarios, besides the requirement of classifying a
key-value sequence accurately, it is also desired to classify a key-value sequence
early, in order to respond fast.
However, these two goals are conflicting in nature, and it is challenging
to achieve them simultaneously.
In this work, we formulate a novel {\em tangled key-value sequence early
  classification} problem, where a tangled key-value sequence is a mixture of
several concurrent key-value sequences with different keys.
The goal is to classify each individual key-value sequence sharing a same key both
accurately and early.
To address this problem, we propose a novel method, i.e., {\em Key-Value sequence 
Early Co-classification} (KVEC), which leverages both inner- and inter-correlations
of items in a tangled key-value sequence through key correlation and value
correlation to learn a better sequence representation.
Meanwhile, a time-aware halting policy decides when to stop the ongoing key-value
sequence and classify it based on current sequence representation.
Experiments on both real-world and synthetic datasets demonstrate that our method 
outperforms the state-of-the-art baselines significantly.
KVEC improves the prediction accuracy by up to $4.7 - 17.5\%$ under the same prediction
earliness condition, and improves the harmonic mean of accuracy and earliness by
up to $3.7 - 14.0\%$.


\end{abstract}



\section{Introduction}
\label{sec:introduction}

The {\em key-value sequence data} has become ubiquitous in the big data era and
such kind of data naturally appears in a wide variety of real-world applications.
For instance, in e-commerce, the customers' product purchasing history data is
often organized as a {\em user-product sequence}, where each item in the sequence
represents a user (i.e., the key) purchasing a particular product (i.e., the
value).
In network traffic analysis, the network packets forwarded by a network device
(such as a router or a switch) can be abstracted as a {\em packet sequence}, and
the packet sequence is further separated into {\em network flows}~\cite{NetFlow},
where the five-tuple in the packet header, i.e., source/destination IP,
source/destination port, and protocol, can be viewed as the key, while the other
fields in the packet header and the payload can be viewed as the value.

\begin{figure}[t]
  \centering
  \input{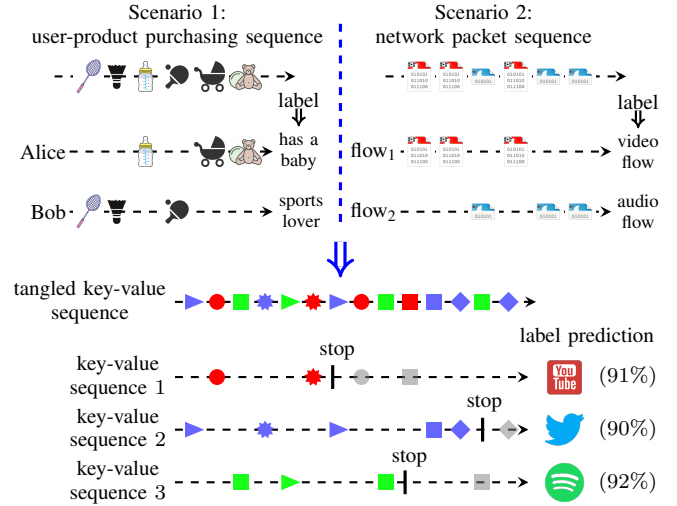}
  \caption{Tangled key-value sequence early classification.
    Items of different shapes represent different network packets.
    Packets of the same color mean that they have the same five-tuple and belong
    to the same network flow.
    We want to classify each network flow both accurately and early.}
  \label{fig:exam}
\end{figure}

A key-value sequence may consist of items having different keys.
While in many applications, it is very helpful to {\em accurately} infer the
labels for each key-value sequence sharing the same key.
For example, in e-commerce, to provide personalized product recommendation
services, it is required to accurately infer a customer's interests, demographics,
etc, based on the customer's purchasing records, aka {\em user
  profiling}~\cite{Yan2020profile}.
In networking management, to detect malicious network traffic and improve the
Quality-of-Service (QoS) of networking, it is important to accurately infer the
application type of each network flow (e.g., a video flow or an audio flow), aka
{\em traffic classification}~\cite{Zheng2020,Rezaei2020}.
In the following discussions, we commonly refer to these tasks as the key-value
sequence classification task, as illustrated in \cref{fig:exam}.

In practice, besides the requirement of classifying a key-value sequence
accurately, it is often better to correctly classify a key-value sequence
\emph{early} than late.
For example, if we are able to accurately obtain a customer's profile just based
on her first few purchasing records, we can quickly learn the interests of a new
customer and recommend proper products to her, which is crucial to attract new
customers and improve the customer retention rate of the platform.
In networking management, if we are able to correctly infer the application type
of a network flow just based on its first few packets, then the router can perform
routing operations timely and better network QoS may be achieved.
However, classifying a key-value sequence both early and accurately are two
conflicting goals in nature.
If we want to classify a key-value sequence accurately, we have to wait to observe
more items in the sequence until enough discriminative features have been collected,
which, however, will
violate the goal of {\em early} classification, and vice versa.
It is thus a highly challenging task to classify a key-value sequence both {\em early}
and {\em accurately}.

In the literature, there have been several efforts aiming to classify a
{\em time series} early and accurately, e.g., electrocardiogram (ECG) time
series early classification~\cite{Huang2022a,Huang2022b},
smart phone sensor data early classification~\cite{Hartvigsen2020}, etc.
We emphasize that due to two major differences between the time series data
considered in previous works and the key-value sequence data considered in our
work, these existing methods are not applicable in our case.

Firstly, time series data consists of consecutive numerical points, where the data
points at consecutive time steps signify the trend of the series.
In the task of time series early classification, the representation model captures
this trend information within each time series to assign labels.
For example, heart disease can be detected by identifying abnormal trends in ECG
time series.
In contrast, within a key-value sequence, each item contains both a key
field, serving as an affiliation indicator, and a value field that ensembles the
semantic information of the item.
In this case, the representation model is required to acquire the latent
semantic information embedded within the key-value sequence.
For example, this involves tasks such as identifying a user's interests from a
user-product sequence or detecting abnormal interaction patterns in
client-server communication through a packet sequence.
Consequently, the representation learning paradigm proposed in previous time
series early classification methods is not suitable for key-value sequence data.
This motivates us to find new methods.

Secondly, in all previous settings of time series early
classification~\cite{Huang2022a,Huang2022b,Hartvigsen2019,Hartvigsen2020}, each
individual time series is considered to be independent with each other.
For example, one patient's ECG time series is indeed independent with another
patient's.
While in our setting, the collected key-value sequence data is actually a mixture
of items with different keys.
Importantly, there may be rich correlations among items both within a
key-value sequence sharing the same key and in key-value sequences with
different keys, and we therefore call it the {\em tangled key-value sequence}
(cf.~\cref{fig:exam}).
Items within an individual key-value sequence sharing the same key are
inherently correlated because they come from a congenetic key.
Moreover, items from different sequences in a tangled key-value sequence may
also exhibit correlations, particularly when there is local similarity among
them.
For example, customers with similar purchasing records may share similar
shopping preferences, and network flows with similar packets may result from the
same attack behavior.
These correlations play a vital role in purifying and enriching the sequence
representation, especially in the early classification setting.
Our goal is to explore these correlations among items in a tangled key-value
sequence to classify each individual key-value sequence sharing the same key both
early and accurately.

In this work, we formulate the {\em tangled key-value sequence early
  classification} problem, and propose a novel solution, namely {\em Key-Value
  sequence Early Co-classification} (KVEC), to address this problem.
The main idea of KVEC is to leverage the correlations among items in a tangled
key-value sequence to learn better sequence representations and hence improve both
prediction accuracy and earliness.
To understand our idea in a high level, consider the e-commerce user profiling
example.
For a new user with few purchasing records, we can leverage similar purchasing
behaviors or popular product combinations from existing users in the platform to
build the new user's profile (which is actually a well-known idea in the
literature of recommendation systems).
While in network traffic analysis, similar packet sequence patterns often imply
similar network functionalities (e.g., similar application type and similar
network anomalies), and these correlations are therefore helpful to identify a
network traffic flow based on its first few packets.

Our proposed solution KVEC mainly consists of two modules, i.e., the {\em
  key-value sequence representation learning} (KVRL) module and the {\em early
  co-classification timing learning} (ECTL) module.
First, the KVRL module learns semantically enhanced sequence representations by
leveraging the correlations of items in a tangled key-value sequence.
Then, the ECTL module adaptively decides whether to continue requesting more items
from the ongoing sequence, or just stop and start performing classification via a
classification neural network.
During model training, these modules are jointly optimized to encourage them to
collaborate with each other and achieve the goal of accurate and early key-value
sequence classification.

Our main contributions can be summarized below:
\begin{itemize}
\item To the best of our knowledge, we are the first to study the general tangled
  key-value sequence early classification problem --- a problem naturally arises
  in many real-world applications, ranging from e-commerce user profiling, to
  network traffic classification (\cref{sec:problem}).

\item We propose a novel solution KVEC to classify the key-value sequences in a
  tangled key-value sequence both early and accurately.
  The key idea of KVEC is to leverage both the key correlation and value
  correlation among items in a tangled key-value sequence to learn expressive
  sequence representations (\cref{sec:method}).

\item We conduct extensive experiments on a variety of datasets (four real-word
  datasets and a synthetic dataset) to validate the effectiveness of KVEC.
  The results show that KVEC improves the prediction accuracy by up to $4.7 -
  17.5\%$ under the same prediction earliness condition, and improves the harmonic
  mean of accuracy and earliness by up to $3.7 - 14.0\%$
  (\cref{sec:experiments}).
\end{itemize}


\section{Related Work}
\label{sec:related_work}

Our work is most related to studies on key-value sequence data mining and time
series early classification, which we summarize as follows.

\header{Key-Value Sequence Data Mining}.
The key-value sequence data is composed of a series of items in
chronological order, where each item is a timestamped key-value pair.
Key-value sequence data mining technologies have wide applications in
recommendation systems and network traffic engineering~\cite{Traffic}.
For example, in e-commerce, analyzing user-product purchasing sequence data can
help build accurate user profiles~\cite{Xu2022profile,Yan2020profile}, which is
important to provide accurate personalized recommendation
services~\cite{Fang2020,Lv2019}.
In network traffic analysis, classifying network packet key-value sequences is
important in identifying application types~\cite{Liu2019,Sirinam2019,Zheng2020},
detecting malicious network intrusions~\cite{Hayes2016,Xu2020}, understanding user
behaviors~\cite{Fu2016,Liu2017}, fingerprinting IoT
devices~\cite{Ma2020iot,Ma2021iot}, etc.

Nevertheless, existing key-value sequence data mining methods all
rely on the complete key-value sequences, and they are sluggish to scenarios
where earliness is critical.
This motivates us to propose this novel problem, i.e., the tangled key-value sequence
early classification problem, where the goal is to achieve both early and
accurate classification of each individual key-value sequence.

\header{Time Series Early Classification}.
The time series data refers to a series of numerical points or vectors
arranged chronologically, representing observations of a single or multiple
variables within a specific time range.
The time series early classification (TSEC) problem focuses on rapidly assigning
the correct label to a time series without the necessity to wait for the
complete series to be observed~\cite{Gupta2020review}.
The TSEC problem finds applications in various domains, including medical
diagnostics~\cite{Hartvigsen2019,Huang2022a,Huang2022b,Hartvigsen2022} and human
activity recognition~\cite{Hartvigsen2020,Gupta2020,Hartvigsen2022}.
According to the strategies for achieving early classification, existing solutions
to the TSEC problem can be categorized into three groups, i.e., feature based
approaches, prefix based approaches, and model based approaches.

Feature based approaches utilize pre-extracted discriminative subseries
as indicators for early classification~\cite{Ye2009,Ghalwash2012,Ghalwash2013}.
Once these indicators are detected in a testing instance, the classifier
immediately performs classification without waiting for further data points
to be observed~\cite{Xing2009,Xing2011,Xing2012,He2015,Schafer2020}.
In contrast, prefix based approaches utilize a set of probabilistic classifiers
stopped at each time step, following a predefined stopping rule to strike a
balance between classification accuracy and earliness~\cite{Mori2016,Mori2017}.
The classifier performs classification only when the current earliness-accuracy
trade-off aligns with the stopping rule.
Due to the absence of joint learning between feature extraction and early
classification, both feature based approaches and prefix based approaches exhibit
poor performance on real-world time series data.

Model-based approaches utilize end-to-end trainable neural networks,
typically combining a time series representation model and a learnable stopping
policy, to address the TSEC problem.
In this framework, the representation of the time series is iteratively learned
in a streaming manner, and a stopping policy then is used to determine the appropriate
early classification position based on the current
representation~\cite{Huang2018,Gupta2020,Russwurm2023,Ebihara2020}.
For instance, in~\cite{Russwurm2019}, a Long Short-Term Memory (LSTM) network
is employed as a feature extractor to capture temporal data point dependencies
and represent them as hidden vectors.
Subsequently, a Convolutional Neural Network (CNN)
functions as the stopping policy to predict the appropriate early classification
position based on these hidden vectors.
Furthermore, in works such
as~\cite{Martinez2018,Martinez2020,Hartvigsen2019,Huang2022a}, researchers
formulate the TSEC problem as the Partially Observable Markov Decision Process
(POMDP), and solve it by introducing a novel halting policy based on
reinforcement learning.
These approaches regard the correctness and time delay of early classification
as rewards and feed them back to the halting policy,
promoting collaboration between the feature extractor
and the halting policy, resulting in state-of-the-art performance in tackling
the TSEC problem~\cite{Hartvigsen2022,Huang2022a,Huang2022b}.

However, as described in Introduction, due to the substantial
difference between time series data and key-value sequence data, existing TSEC
methods are not suitable for the tangled key-value sequence early classification
problem.
Motivated by these challenges, we propose a novel method (i.e., KVEC) to address
this problem.
The core idea of KVEC is a novel representation learning approach that comprehensively
explores correlations within tangled key-value sequences to obtain semantically
enhanced sequence representations,
resulting in superior early classification performance.

\section{Notations and Problem Formulation}
\label{sec:problem}

In general, we denote a {\em tangled key-value sequence} by
$\CS\triangleq\left(\pr{k,\bv}\colon k\in\CK, \bv\in\CV_1\times \cdots\times\CV_l\right)$,
where each item $\pr{k,\bv}$ consists of two fields, i.e., the \emph{key} field
$k$, and the \emph{value} field $\bv$.
Here, the key space $\CK$ is a finite set, representing all keys in the sequence,
and the value $\bv$ is an $l$-dimensional vector where the $i$-th dimension is from
space $\CV_i$ for $1\leq i\leq l$.
We assume items in the sequence arrive sequentially, one at a time.
For the ease of presentation, we will use $e\in\CS$ to denote an item, and use
$e.k$ and $e.\bv$ to refer to its key and value, respectively.

As mentioned in Introduction, the tangled key-value sequence data can model a wide
range of real-world applications.
For example, in a user-product purchasing sequence, the key may be the user ID,
and the value may be the attribute vector of the product purchased by the user.
In a network packet sequence, the key of a network packet may be the five-tuple
extracted from the packet header, and the value may be the extracted feature
vector (either from the header or payload), e.g., protocol, port number, TTL,
payload size, etc.

Furthermore, let $\CS_k\triangleq\left(\pr{k,\bv}\colon\pr{k,\bv}\in\CS\right)$ denote a
key-value sequence sharing a same key $k\in\CK$ in $\CS$.
The relation between a tangled key-value sequence $\CS$ and each key-value
sequence $\CS_k\subseteq\CS$ is illustrated in \cref{fig:exam}.
For convenience, we will use $|\CS_k|$ to denote the length of sequence $\CS_k$.

In many applications, we are interested in assigning a label (or multiple labels)
to $\CS_k$.
For example, in e-commerce, $\CS_k$ may represent the purchasing records of a
specific user $k$, and the user profiling task can be formulated as inferring
the labels of user $k$ based on his key-value sequence $\CS_k$.
In networking management, $\CS_k$ may represent a network flow consisting of
packets sharing a same five-tuple $k$, and in many cases, we need to assign an
application type to $\CS_k$ in order to detect malicious traffic, optimize packets
routing, etc.

As we mentioned earlier, besides classifying a key-value sequence $\CS_k$
accurately, it is also desired to classify each key-value sequence $\CS_k$ as
early as possible, which could greatly benefit the recommendation system and
improve network QoS.
As items in the tangled key-value sequence keep arriving, our goal in this work is
to classify each individual key-value sequence $\CS_k$ both early and accurately,
based on currently {\em observed} items in $\CS$.
We refer to this problem as the {\em tangled key-value sequence early
  classification problem}, or just the {\em early classification problem} if no
confusion arises.



\section{Key-Value Sequence Early Co-Classification}
\label{sec:method}

To achieve early classification, a straightforward approach is that we just
collect a few number of items for each key-value sequence for classification, say,
the first $n$ items of each key-value sequence.
There are several drawbacks about this naive approach.
First, it is difficult to find a proper universal $n$ for each key-value sequence,
and we have to costly enumerate every possible $n$ to find the optimal one.
Second, even though $n$ can be determined, a universal $n$ for each key-value
sequence may be not appropriate.
Because some sequences may be ``easy'' to classify, and we just need to collect a
fewer number of items than the fixed universal $n$; while some sequences may be
``hard''  to classify, and we have to collect more items than the fixed universal
$n$.
Obviously, a fixed universal $n$ will harm earliness (accuracy) for easy (hard)
sequences.
To address the issues of this naive approach, we present a better solution in this
section, namely {\em Key-Value sequence Early Co-classification} (KVEC).

\subsection{Overview of KVEC}
\label{ss:overview}

Instead, we propose that each key-value sequence should be collected for a
different number of items for observation and investigation. Let $n_k$ be the
least number of items that should be collected from sequence $\CS_k$.
Hence, the number of observations $n_k$ plays a crucial role in balancing the
prediction earliness and accuracy for sequence $\CS_k$.
On the one hand, we want the number of observations for a sequence to be small
enough so that we can predict the sequence's labels early.
On the other hand, the number of observations should be large enough so that we
can collect enough items from the sequence and predict the sequence's labels
accurately.
Therefore, the tangled key-value sequence early classification problem boils down
to how to adaptively determine the number of observations $n_k$ for each key-value
sequence $\CS_k$.

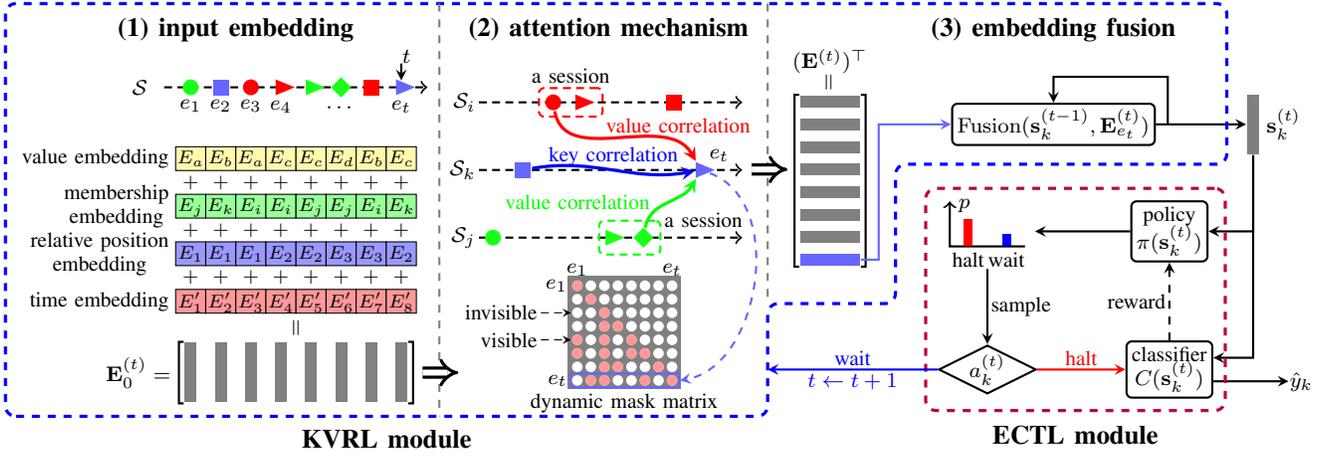
\begin{figure*}[htp]
  \centering
  \begin{tikzpicture}[thick, font=\footnotesize, >=stealth,
  every node/.style={inner sep=0pt},
  cir/.style={circle,minimum size=6pt},
  rec/.style={rectangle,minimum size=6pt},
  tri/.style={isosceles triangle,minimum size=6pt},
  dia/.style={diamond,minimum size=8pt},
  msk/.style={circle,minimum size=4pt,fill=white},
  c1/.style={fill=red},
  c2/.style={fill=blue!60},
  c3/.style={fill=green!90},
  box/.style={draw,rectangle,thin,minimum width=4mm, minimum height=2ex,fill=#1!40,
    font=\scriptsize,text depth=1pt},
  stxt/.style={font=\footnotesize,align=center,inner sep=1pt},
  session/.style={draw,rectangle,densely dashed,inner sep=2pt,rounded corners=2pt},
  arr/.style={densely dashed,-Classical TikZ Rightarrow},
  cor/.style={->,very thick},
  vec/.style={fill=gray,rectangle,minimum width=1ex,minimum height=5ex},
  row_vec/.style={fill=gray,rectangle,minimum width=5ex,minimum height=1ex},
  fun/.style={draw,rectangle,rounded corners=2pt,inner sep=2pt,align=center},
  fitbox/.style={draw=gray,very thick,dashed,inner sep=1ex,rounded corners=4pt},
  ]

  \begin{scope}
    \coordinate (O);
    \draw[arr] (O) ++(.3ex,0) -- ++(3.5,0) coordinate (right_axis);
    \foreach \nd/\c [count=\i] in {cir/3,rec/2,cir/1,tri/1,tri/3,dia/3,rec/1,tri/2}{
      \coordinate[right=\i*.4 of O] (C\i);
      \node[\nd,c\c] (n\i) at (C\i) {};
    }
    \node[left=1ex of O] (t) {$\CS$};
    \draw[<-] (n8) -- ++(0,2ex) node[anchor=south west] (t1) {$t$};
    \node[stxt,below=0 of n1] {$e_1$};
    \node[stxt,below=0 of n2] {$e_2$};
    \node[stxt,below=0 of n3] {$e_3$};
    \node[stxt,below=1pt of n4] {$e_4$};
    \node[stxt,below=0 of n6] {$\cdots$};
    \node[stxt,below=1pt of n8] {$e_t$};

    \coordinate[below=6ex of O] (P1);
    \coordinate[below=2ex of P1] (P2);
    \coordinate[below=2ex of P2] (P3);
    \coordinate[below=2ex of P3] (P4);
    \coordinate[below=2ex of P4] (P5);
    \coordinate[below=2ex of P5] (P6);
    \coordinate[below=2ex of P6] (P7);
    \coordinate[below=6ex of P7] (P8);
    \foreach \v/\m/\r [count=\i] in {a/j/1,b/k/1,a/i/1,c/i/2,c/j/2,d/j/3,b/i/3,c/k/2}{
      \node[box=yellow] at (P1 -| n\i) {$E_\v$};
      \node[box=green] at (P3 -| n\i) {$E_\m$};
      \node[box=blue] at (P5 -| n\i) {$E_\r$};
      \node[box=red] at (P7 -| n\i) (E\i) {$E'_\i$};
      \node[font=\scriptsize] at (P2 -| n\i) {$+$};
      \node[font=\scriptsize] at (P4 -| n\i) {$+$};
      \node[font=\scriptsize] at (P6 -| n\i) {$+$};

      \node[vec] at (P8 -| n\i) (vec\i) {};
    }

    \node[stxt,left=-1ex of P1] (v_embd) {value embedding};
    \node[stxt,left=-1ex of P3] {membership\\embedding};
    \node[stxt,left=-1ex of P5] {relative position\\embedding};
    \node[stxt,left=-1ex of P7] {time embedding};
    \node[above right=.6 and 1 of O,anchor=south,font=\bfseries] (t_embd) {(1) input embedding};

    \node[below right=2ex and .15 of E4.south,rotate=90] {$=$};

    \draw (vec1.north west) -- ++(-.4ex,0) |- (vec1.south west);
    \draw (vec8.north east) -- ++(.4ex,0) |- (vec8.south east);

    \node[left=1ex of vec1] {$\bE_0^{(t)}=$};
  \end{scope}

  \begin{scope}[xshift=4.2cm]
    \coordinate (O);
    \foreach \nd [count=\i] in {cir,rec,cir,tri,tri,cir,rec,tri}{
      \node[\nd,right=\i*.4 - .3 of O] (d\i) {};
    }

    \coordinate[below=.2 of O] (O1);
    \draw[arr] (O1) ++(.2ex,0) -- ++(3.5,0);
    \foreach \nd/\i [count=\j] in {cir/3,tri/4,rec/7}{\node[\nd,c1] (S1_\j) at (O1 -| d\i) {};}

    \coordinate[below=.9 of O1] (O2);
    \draw[arr] (O2) ++(.2ex,0) -- ++(3.5,0);
    \foreach \nd/\i [count=\j] in {rec/2,tri/8}{\node[\nd,c2] (S2_\j) at (O2 -| d\i) {};}
    \node[stxt,above right=0 of S2_2] {$e_t$};

    \coordinate[below=.9 of O2] (O3);
    \draw[arr] (O3) ++(.2ex,0) -- ++(3.5,0);
    \foreach \nd/\i [count=\j] in {cir/1,tri/5,dia/6}{\node[\nd,c3] (S3_\j) at (O3 -| d\i) {};}

    \node[left=0 of O1] (t1) {$\CS_i$};
    \node[left=0 of O2] (t2) {$\CS_k$};
    \node[left=0 of O3] (t3) {$\CS_j$};

    \node[right=1.5 of t_embd,font=\bfseries] {(2) attention mechanism};

    \node[session,red,fit={(S1_1) (S1_2)}] (s1) {};
    \node[session,green,fit={(S3_2) (S3_3)}] (s2) {};
    \node[stxt,above=0 of s1] {a session};
    \node[stxt,right=0 of s2.30] {a session};

    \draw[cor,red] (S1_1) to [out=-80,in=120] node[stxt,above=3pt,pos=.85] (val_cor) {value correlation} (S2_2);
    \draw[cor,green] (S3_3) to [out=80,in=-120] node[stxt,left=5pt] (e) {value correlation} (S2_2);
    \draw[cor,blue] (S2_1) to [out=10,in=190] node[stxt,above] {key correlation} (S2_2);

    \node[rectangle,fill=gray,minimum size=1.5cm, below right=.5 and 1.2 of O3] (M) {};
    \foreach \i in {1,...,8} {
      \foreach \j in {1,...,8} {
        \node[msk,below right=\i*0.18-0.1 and \j*0.18-0.1 of M.north west] (m\i\j) {};
      }
      \node[msk,fill=red!40] at (m\i\i) {};
    }
    \foreach \i/\j in {4/3,5/1,5/3,6/1,6/3,6/5,7/3,7/4,8/2,8/3,8/6}{
      \node[msk,fill=red!40] at (m\i\j) {};
    }
    \node[stxt,left=0 of m11] {$e_1$};
    \node[stxt,left=0 of m81] {$e_t$};
    \node[stxt,above=0 of m11] {$e_1$};
    \node[stxt,above=0 of m18] {$e_t$};
    \node[stxt,left=.4 of m31] (invisible) {invisible};
    \node[stxt,left=.4 of m51] (visible) {visible};
    \draw[densely dashed,thin,->] (invisible) -- (m31);
    \draw[densely dashed,thin,->] (visible) -- (m51);

    \node[draw=blue!60,fit={(m81) (m88)},inner sep=.2pt] (row8) {};
    \draw[draw=blue!60,->,dashed,] (S2_2) to [out=-20,in=10] (row8.east);

    \node[stxt,below=0 of M] {dynamic mask matrix};

    \coordinate[above right= .2 and 4.7 of O] (O4);
    \foreach \i in {1,...,8}{
      \node[row_vec, below=.3*\i of O4] (rvec\i) {};
    }
    \node[row_vec,fill=blue!60] at (rvec8) {};

    \draw (rvec1.north west) -- ++(-.4ex,0) |- (rvec8.south west);
    \draw (rvec1.north east) -- ++(.4ex,0) |- (rvec8.south east);
    \node[stxt,above right=0 and .1 of rvec1.north,rotate=90] {$=$};
    \node[stxt,above=1.5ex of rvec1] {$(\bE^{(t)})^\top$};

    \node[fun,right=1.2 of rvec2] (fusion) {$\mathrm{Fusion}(\bs_k^{(t-1)},\bE^{(t)}_{e_t})$};
    \node[font=\bfseries] at (t_embd -| fusion) {(3) embedding fusion};

    \draw[->] (fusion.east) ++(1ex,0) -- ++(0,4ex) -| (fusion);
    \draw[->,blue!60] (rvec8.east) -- ++(1.5ex,0) |- (fusion);
  \end{scope}

  \node[vec,right=1.2 of fusion] (s) {};
  \node[right=2pt of s] {$\bs_k^{(t)}$};
  \draw[->] (fusion) -- (s);

  \node[fun,below left=.6 and .5 of s] (policy) {policy\\$\pi(\bs_k^{(t)})$};

  \coordinate[below left=.2 and 1.9 of policy.west] (X);
  \draw[-Classical TikZ Rightarrow] (X) ++(.5,0) -- ++(-1,0) -- ++(0,.6) node[stxt,right=2pt,pos=.9]{$p$};
  \node[fill=red,minimum width=.7ex,minimum height=2.2ex,left=1.6ex of X,anchor=south] (bar1) {};
  \node[fill=blue,minimum width=.7ex,minimum height=1ex,right=1.6ex of X,anchor=south] (bar2) {};
  \node[stxt,below=1pt of bar1] {halt};
  \node[stxt,below=1pt of bar2] {wait};

  \node[draw,shape aspect=2,diamond,below=1.3 of X] (action) {$a_k^{(t)}$};
  \node[fun] (classifier) at (policy |- action) {classifier\\$C(\bs_k^{(t)})$};
  \node[right= 1 of classifier.-15] (yk) {$\hat{y}_k$};

  \draw[->] (s) |- (policy.east);
  \draw[->] (s) |- (classifier.15);
  \draw[<-] (policy -| X) ++(4ex,0) -- (policy);
  \draw[->] (X) ++(0,-.3) -- node[stxt,right] {sample} (action);
  \draw[->,red] (action) -- node[stxt,above] {halt} (classifier);
  \draw[->,dashed] (classifier) -- node[stxt,left] {reward} (policy);
  \draw[->] (classifier.east |- yk) -- (yk);

  \node[fitbox,purple,fit={(classifier) (action) (policy)}] (tl_box) {};
  \node[below=1ex of tl_box,font=\bfseries] (bot_rep) {ECTL module};

  \node[inner sep=1ex,fit={(v_embd) (vec8) (t_embd) (val_cor)}] (rep_box) {};
  \node[below=1ex of rep_box,font=\bfseries] (bot_rep) {KVRL module};

  \coordinate[above right=1.5 and 1.7 of rep_box.south east] (BR1);
  \coordinate[above=.3 of tl_box.north east] (BR2);

  \draw[fitbox,blue] (rep_box.south west) -- (rep_box.south east) |- (BR1) |- (BR2) |- (rep_box.north west) -- cycle;

  \draw[->,blue] (action) -- node[stxt,above] {wait}
    node[stxt,below] {$t\gets t+1$} (rep_box.south east |- action);

  \coordinate[right=1ex of right_axis] (RL);
  \draw[fitbox,thick] (rep_box.north -| RL) -- (rep_box.south -| RL);
  \draw[fitbox,thick] (rep_box.north east) -- (rep_box.south east |- BR1);

  \draw[double,very thick,double distance=.5ex,Implies-] (rvec4.west) ++(-1ex,0) -- ++(-3ex,0);
  \draw[double,very thick,double distance=.5ex,-Implies] (vec8.east) ++(1.2ex,0) -- ++(3ex,0);
\end{tikzpicture}
  \caption{Overview of the KVEC framework.}
  \label{fig:workflow}
\end{figure*}

We propose KVEC to resolve this conflict of adaptively determining the number
of observations $n_k$ for each key-value sequence $\CS_k$.
KVEC is armed with two modules, i.e., the {\em Key-Value sequence Representation
  Learning} (KVRL) module, and the {\em Early Co-classification Timing Learning}
(ECTL) module, as illustrated in \cref{fig:workflow}.
The KVRL module is designed to learn an informative representation of the
partially observed key-value sequence by exploiting rich item correlations in the
tangled key-value sequence.
A good sequence representation is the key to improve both prediction earliness and
accuracy.
The ECTL module is designed to adaptively learn to determine a proper number of
observations for each key-value sequence and balance the prediction earliness and
accuracy.

In what follows, we elaborate on each module in proposed KVEC.


\subsection{Key-Value Sequence Representation Learning (KVRL)}
\label{ss:KVRL}

The goal of the KVRL module is to learn an informative representation for each
key-value sequence, and it is the core of KVEC because a good sequence
representation is the key to achieve both early and accurate prediction.
The challenge is that, to achieve early classification, we must collect as few
items as possible for each key-value sequence, which, however, will definitely
harm the sequence representation learning due to data scarcity.

We propose KVRL to address this challenge.
KVRL consists of three major operations to learn a semantic representation of each
key-value sequence, i.e., {\em input embedding}, {\em attention mechanism}, and
{\em embedding fusion}.

\header{Input Embedding}.
The purpose of input embedding is to represent each item $\pr{k,v}\in\CS$ as a
preliminary hidden vector, we consider the semantics embedded in the key field and
value field, respectively.

The value field contains the core semantics of an item.
For example, in the user-product purchasing sequence, the product field provides
the majority of information to understand the $\pr{\text{user},\text{product}}$
item.
Thus, we assign to each item a learned {\em value embedding} corresponding to the
value field, which could be initialized by a pre-trained model and then fine-tuned
later, or completely learned from the provided training data.
As illustrated in the left part of \cref{fig:workflow}, $e_1$ and $e_3$ represent
two items having the same value field (indicated by the same shape), thus their
value embeddings are both represented by a value embedding vector $E_a$.

The key field mainly tells which key-value sequence the item belongs to, and the
position of this item in the corresponding key-value sequence.
Thus, we add a learned {\em membership embedding} to every item indicating its
affiliation relationship with a key-value sequence.
For example, in the left part of \cref{fig:workflow}, because items $e_3$ and
$e_4$ belong to the same key-value sequence (indicated by the same color), their
membership embeddings are both represented by a membership embedding vector $E_i$.
We also add a learned {\em relative position embedding} to every item to indicate
its relative position in a key-value sequence.
For example, in the left part of \cref{fig:workflow}, because $e_3$ and $e_4$ are
the first and second items in the same key-value sequence, their relative position
embeddings are denoted by $E_1$ and $E_2$, respectively.

In addition, in order to make use of the time (or order) information, we also add
a learned {\em time embedding} to every item indicating its arrival time (or
order).

Finally, the input embedding of each item is the sum of its value embedding,
membership embedding, relative position embedding, and time embedding.
Putting all embeddings together, we obtain a dynamic embedding matrix
$\bE_0^{(t)}\in\Real^{d\times t}$, as illustrated in the left part of
\cref{fig:workflow}.
The $i$-th column of the dynamic embedding matrix $\bE_0^{(t)}$ is the
$d$-dimensional embedding vector of the $i$-th item in $\CS$.
As items keep arriving, more columns will be appended to $\bE_0^{(t)}$.




\header{Attention Mechanism}.
In the previous step, each item in a key-value sequence is treated independently.
In practice, items are often correlated, and this correlation can be used to
obtain better item representations especially when only a few items in a key-value
sequence are collected.
In what follows, we consider two types of item correlations in a tangled key-value
sequence, i.e., {\em key correlation} and {\em value correlation}.

The key correlation captures the relation of items that are in a key-value
sequence sharing the same key.
Intuitively, items belonging to a same key-value sequence are more likely to be
related with each other.
For example, a person's interests is usually consistent in a short time period and
hence the person will watch movies of similar genres.
In other words, movies watched by a same person are likely to be related.
Therefore, previous items can guide the representation learning of the current
item in the same key-value sequence.
Formally, we say that two items $e$ and $e'$ are {\em correlated through key
  correlation} if $e.k=e'.k$, denoted by $e\stackrel{\text{\tiny key}}{\sim} e'$.

The value correlation captures the relation of items due to the correlation of
their value fields.
Intuitively, a set of consecutive and time-adjacent items in a key-value sequence
are likely to be correlated.
For example, people often buy combinations of correlated products in a shopping
transaction, which is a well known phenomenon in market basket
analysis~\cite{Guo2019session,Choi2021session}.
In network traffic engineering, it is observed that a set of consecutive and
time-adjacent network packets form a \emph{burst}~\cite{Shen2021,Ma2021iot}, and
network packets in a burst are often related to a fine-grained action such as
logging into an APP and sending a message.
Motivated by these observations, we can build relations among items through their
value fields.

Formally, we define a set of consecutive and time-adjacent items in a key-value
sequence as a \emph{session}.
Items in a session have the same value in a specific subspace $\CV_i$ of the 
value field $\bv$ (e.g., the transmission direction of packet in a packet sequence,
the category of product in a user-product sequence),
and are uninterrupted in time,
as illustrated in the middle part of \cref{fig:workflow}.

We say that two items $e$ and $e'$ are {\em correlated through value correlation}
if their exists a key $k\in\CK$ such that items $\pr{k,e.\bv}$ and $\pr{k,e'.\bv}$
are in the same session of $\CS_k$, denoted by $e\stackrel{\text{\tiny
    value}}{\sim} e'$.
For example, in \cref{fig:workflow}, we have $e_t\stackrel{\text{\tiny
    value}}{\sim} e_3$, because if we change $e_t.k$ to $e_3.k$, then they belong
to a same session in sequence $\CS_i$.


To incorporate key correlation and value correlation in the key-value sequence
representation learning, we design a novel attention mechanism.
First, we define a {\em dynamic mask matrix} to indicate whether there is a
correlation between two items in the tangled key-value sequence, denoted by
$\bM^{(t)}\in\{0,-\infty\}^{t\times t}$.
The $ij$-th element of $\bM^{(t)}$ is given by
\[
  \bM_{ij}^{(t)} \triangleq
  \begin{cases}
    0,       & i=j, \\
    0,       & (e_i\stackrel{\text{\tiny key}}{\sim} e_j\text{ or }
               e_i\stackrel{\text{\tiny value}}{\sim} e_j)\text{ and } j \leq i, \\
    -\infty, & \text{otherwise}.
  \end{cases}
\]
Here, $\bM_{ij}^{(t)}=0$ indicates that items $e_i$ and $e_j$ are correlated,
either through a key or value correlation.
The additional condition $j\leq i$ ensures that $e_i$ can only have correlations
with $e_j$ arrived earlier, thereby guaranteeing their causality.
$\bM_{ij}^{(t)}=-\infty$ indicates that items $e_i$ and $e_j$ are not correlated.
Equivalently, we say that item $e_j$ is visible to item $e_i$ if
$\bM_{ij}^{(t)}=0$; otherwise, $e_j$ is invisible to $e_i$.
An example of the dynamic mask matrix is illustrated in the middle part of
\cref{fig:workflow}.

Next, we linearly project the dynamic embedding matrix $\bE_0^{(t)}$ to the query,
key, and value matrices, i.e., $\bQ, \bK, \bV\in\Real^{d\times t}$, through
learnable projection matrices $\bW_q, \bW_k, \bW_v\in\Real^{d\times d}$,
respectively, i.e.,
\[
  \bQ = \bW_q\bE_0^{(t)},\quad
  \bK = \bW_k\bE_0^{(t)}, \quad
  \bV = \bW_v\bE_0^{(t)}.
\]
We modify the original self-attention mechanism~\cite{Vaswani2017a} by adding the
dynamic mask matrix $\bM^{(t)}$ to make use of the key and value correlations in
the tangled key-value sequence and output a refined embedding matrix
$\bar\bE^{(t)}\in\Real^{d\times t}$, i.e.,
\[
  \bar\bE^{(t)}
  = \mathrm{Attention}(\bQ,\bK,\bV)
  = \softmax{\frac{\bQ^\top\bK + \bM^{(t)}}{\sqrt{d}}}\bV^\top.
\]

Finally, we employ a two-layer fully connected feed forward network to introduce
non-linearity in items' embeddings, and denote the output embedding matrix by
$\bE^{(t)}\in\Real^{d\times t}$, i.e.,
\[
  \bE^{(t)}_e = \mathrm{FFN}(\bar\bE^{(t)}_e)
  = \bW_2\cdot\mathrm{ReLu}(\bW_1\bar\bE^{(t)}_e + \bb_1) + \bb_2
\]
for $e\in\CS$.
Here, $\bW_1\in\Real^{d \times d'}$, $\bW_2 \in \Real^{d' \times d}$,
$\bb_1\in\Real^{d'}$, $\bb_2 \in \Real^d$ are learnable parameters.
$\bE^{(t)}_e$ and $\bar\bE^{(t)}_e$ denote the vectors corresponding to
item $e$ in the matrices $\bE^{(t)}$ and $\bar\bE^{(t)}$, respectively.

\header{Embedding Fusion}.
Up till now, for each key-value sequence $\CS_k$ sharing the key $k$, we have
obtained the item embedding $\bE^{(t)}_e$ for every $e\in\CS_k$.
We still need to fuse these item embeddings to obtain the final representation
vector for sequence $\CS_k$, denoted by $\bs_k^{(t)}$.
To this end, we propose the embedding fusion operation.

A straightforward way for embedding fusion is that we can simply combine item
embeddings using addition, concatenation, or averaging operations.
However, we find that these parameter-free operations often result in poor
prediction results due to noise aggregation when handling the key-value sequence
data.
Instead, we propose a learning method for embedding fusion inspired by the success
of LSTM~\cite{Hochreiter1997} for handling sequence data.
We employ a multiple gating mechanism as in LSTM to selectively filter noise
information, fuse item embeddings, and generate the sequence representation, as
illustrated in the top-right part of \cref{fig:workflow}.

Formally, at time $t$, the current sequence representation $\bs_k^{(t)}$ is a
function of the previous sequence representation $\bs_k^{(t-1)}$ and the current
item representation $\bE^{(t)}_e$, i.e.,
\[
  \bs_k^{(t)} = \mathrm{Fusion}(\bs_k^{(t-1)}, \bE^{(t)}_e).
\]
Here, the $\mathrm{Fusion}$ operation is implemented in a way similar to LSTM but
dedicatedly adapted to our case.
We first define three gates, i.e.,
\begin{align*}
  f_t &= \sigma(\bW_f[\bs_k^{(t-1)};\bE^{(t)}_e]+\bb_f), \\
  i_t &= \sigma(\bW_i[\bs_k^{(t-1)};\bE^{(t)}_e]+\bb_i), \\
  o_t &= \sigma(\bW_o[\bs_k^{(t-1)};\bE^{(t)}_e]+\bb_o),
\end{align*}
where $f_t$, $i_t$, and $o_t$ denote the forget gate, input gate, and output gate,
respectively; $\bW_f$, $\bW_i$, $\bW_o$, $\bb_f$, $\bb_i$, and $\bb_o$ are
learnable weight matrices and bias vectors; $\sigma$ denotes the sigmoid function;
$[;]$ denotes vector concatenation.
The forget gate $f_t$ determines which irrelevant information should be removed
from the cell memory, and the input gate $i_t$ controls the addition of new item
representation to the cell memory, i.e.,
\[
  C_t = f_t\odot C_{t-1} + i_t\odot\eta(\bW_c[\bs_k^{(t-1)};\bE^{(t)}_e]+\bb_c)
\]
where $C_t$ denotes the cell memory at time $t$; $\bW_c$ and $\bb_c$ are learnable
weight matrix and bias vector; $\odot$ is the Hadamard product; $\eta(\cdot)$
denotes the hyperbolic tangent function.
Finally, the sequence representation $\bs_k^{(t)}\in\Real^d$ is computed by
applying the output gate to the non-linear transformation of cell memory $C_t$,
i.e.,
\[
  \bs_k^{(t)} = o_t\odot\eta(C_t).
\]

\subsection{Early Co-classification Timing Learning (ECTL)}

Given the current representation vector $\bs_k^{(t)}$ of key-value sequence
$\CS_k$, the ECTL module then adaptively decides whether enough items in $\CS_k$
have been collected and the sequence is ready for classification, or more items in
$\CS_k$ need to be collected to keep updating $\bs_k^{(t)}$.
Therefore, the ECTL module is the key decision component to control the workflow
in our KVEC framework.
We adopt the reinforcement learning technique to implement ECTL.

At each time step, the ECTL module takes the sequence representation $\bs_k^{(t)}$
as the environment state, chooses an action according to a learned policy network,
and a reward is gained according to the chosen action.
We train the policy network in ECTL to achieve the maximum accumulated reward over
time.
The structure of ECTL is illustrated in the bottom-right part of
\cref{fig:workflow}.


\header{State}.
In reinforcement learning, states represent the current environment, and the agent
takes actions according to current state.
In our case, we choose the key-value sequence representation vector $\bs_k^{(t)}$
at time step $t$ as the state.

\header{Policy}.
The agent leverages a policy to decide the next action according to the current
environment state.
Let $\pi\colon\Real^d\mapsto[0,1]$ be the policy function that takes a state as
the input and outputs a probability which will be used to decide the next action.
Following existing works~\cite{Hartvigsen2019,Martinez2020}, we use a neural
network to approximate the policy function $\pi$, i.e.,
\[
  \pi(\bs_k^{(t)}) = \sigma(\bw_\pi\cdot\bs_k^{(t)} + b_\pi)
\]
where $\bw_\pi\in\Real^d$ and $b_\pi\in\Real$ are learned parameters.

\header{Action}.
The agent chooses an action $a_k^{(t)}$ for $\CS_k$ from an action space
$\CA\triangleq\{\mathit{Halt},\mathit{Wait}\}$ according to the policy defined
previously, and
\begin{align*}
  P(a_k^{(t)}=\mathit{Halt}) &= \pi(\bs_k^{(t)}),\\
  P(a_k^{(t)}=\mathit{Wait}) &= 1- \pi(\bs_k^{(t)}).
\end{align*}
If action $a_k^{(t)} = \mathit{Halt}$, the current sequence $\CS_k$ is halted,
which means that enough items from $\CS_k$ have been collected and $\bs_k^{(t)}$
is ready to be delivered to the classification module for sequence classification;
otherwise, the action $\mathit{Wait}$ means that we need to keep collecting more
items in $\CS_k$ to update its representation $\bs_k^{(t)}$.

\header{Reward}.
In order to steer the agent to optimize towards the expected direction, the
classifier returns the correctness of predictions as rewards to evaluate the
performance of policy.
When the classifier assigns a correct label for the sequence representation
$\bs_k^{(t)}$, the ECTL module receives a positive reward $r_k^{(t)}=1$;
otherwise, it receives a negative reward $r_k^{(t)}=-1$.

\subsection{Classification Network}

Once the ECTL module chooses an action $\mathit{Halt}$ for key-value sequence
$\CS_k$, the classifier needs to determine the sequence label according to the
sequence's current representation $\bs_k^{(t)}$.
Note that because we need to classify different key-value sequences sequentially,
this classification problem is slightly different from the traditional multi-class
classification problem that treats each example independently.


We implement the classifier by a neural network consisting of a fully-connected
layer followed by a softmax layer.
When $\CS_k$ is halted, the classification network takes its representation
$\bs_k^{(t)}$ as the input, and outputs a probability distribution $\bp_k\in
\Real^C$ over $C$ labels.
Formally,
\[
  \bp_k = \softmax{\bW_c\bs_k^{(t)} + \bb_c}
\]
where $\bW_c\in\Real^{C\times d}$ and $\bb_c\in\Real^C$ are learned parameters.
Finally, the sequence label for $\CS_k$ is indicated by the dimension with the
maximum value in $\bp_k$, i.e., $\hat{y}_k = \arg\max_{i}\bp_{k,i}$.

\subsection{Model Training}
\label{ss:training}

In model training, our goal is to jointly minimize the prediction error of the
classification network, maximize the accumulate reward gained by the policy
network in ECTL, and also encourage early classification in ECTL.

The first part of the optimization goal is focused on the sequence representation
learning, i.e., the KVRL module, as well as the classification network.
Let $\btheta_1$ denote the parameters related to these two modules.
We use the cross entropy to evaluate the loss of prediction error, i.e.,
\[
  l_1(\btheta_1) \triangleq
  -\sum_{k=1}^K\sum_{c=1}^C\indr{y_k=c}\log\bp_{k,c}(\btheta_1)
\]
where $K$ is the number of key-value sequences in a tangled key-value sequence
$\CS$, $y_k\in\{1,\ldots,C\}$ is the ground-truth label of $\CS_k$, and
$\indr{\cdot}$ denotes the indicator function.


The second part of the goal is focused on the ECTL module.
Let $\btheta_\pi$ denote the parameters of the halting policy network in ECTL.
Our goal is to maximize the expected cumulative reward
$\E[\pi]{\sum_{i=1}^{n_k}r_k^{(i)}}$ for each sequence $\CS_k$.\footnote{ Because
  the item arrived at time $t$ is not always an item belonging to sequence $\CS_k$
  for some given $k$, we thus use $r_k^{(i)}$ to denote the reward gained after
  the agent takes an action on the $i$-th item in $\CS_k$.
  Similarly, $a_k^{(i)}$ is the action the agent taking on the $i$-th item in
  $\CS_k$, and $\bs_k^{(i)}$ is the representation of sequence $\CS_k$ after
  observing its $i$-th item.
  Note that given $\CS_k$, we have $1\leq i\leq n_k$.}
However, due to the non-differentiable nature of action sampling in ECTL, directly
optimizing the policy network using gradient descent is difficult.
Instead, following the classic REINFORCE with Baseline method~\cite{Sutton2020book}
we transform the original target to a surrogate loss function,
i.e.,
\[
  l_2(\btheta_2) \triangleq
  -\sum_{k=1}^K\sum_{i=1}^{n_k}
  \big(R_k^{(i)}-b_k^{(i)}(\btheta_b)\big)
  \log P(a_k^{(i)}|\bs_k^{(i)};\btheta_\pi),
\]
where $R_k^{(i)}\triangleq\sum_{s=i+1}^{n_k}r_k^{(s)}$ is the cumulative reward
obtained from the execution of action $a_k^{(i)}$ until the end of an episode of
sequence $\CS_k$, $b_k^{(i)}(\btheta_b)$ denotes a baseline state-value function
of current state $\bs_k^{(i)}$ with parameters $\btheta_b$, and
$\btheta_2\triangleq (\btheta_\pi, \btheta_b)$ are the parameters.
The purpose of maximizing the cumulative reward with respect to a baseline instead
of only maximizing the cumulative reward has the advantage of reducing the
variance and hence speeding the convergence of learning (cf.~\cite{Sutton2020book} for
more details).
We implement the baseline by a shallow feed-forward neural network.


The third part of the optimization goal is also related to the ECTL module, and it
is used to penalize late classification.
We notice that, without this penalty, ECTL will tend to take the correctness of
predictions as a core objective, and delay the $\mathrm{Halt}$ action to ensure
prediction correctness.
As a result, prediction earliness is not guaranteed.
Therefore, to encourage early prediction, we introduce the following third loss,
i.e.,
\[
  l_3(\btheta_3) \triangleq -\sum_{k=1}^K\sum_{i=1}^{n_k}
  \log P(a_k^{(i)} = \mathrm{Halt}|\bs_k^{(i)};\btheta_\pi),
\]
where $\btheta_3=\btheta_\pi$.
This loss enforces the halting policy to learn a larger halting probability
$P(a_k^{(i)}=\mathrm{Halt}|\bs_k^{(i)})$ and hence make a prediction early.

The ideal approach for model training is to update the parameters of KVRL and ECTL
independently using their respective supervisory signals (i.e., predicted labels
for KVRL and rewards for halting policy).
However, a challenging issue in early classification is the absence of labeled
exact halting positions, making it difficult to accurately measure the correctness
of actions taken by the halting policy.
In KVEC, the correctness of actions is assessed using the predicted labels from
the classification network.
If the key-value sequence is correctly classified, all actions taken by the
halting policy, including {\em Wait} and {\em Halt}, are deemed correct and return
positive rewards, and vice versa.
This coupling of KVRL and ECTL necessitates considering their parameters as a
whole and optimizing them synchronously.
Therefore, combining above three losses, we obtain the total training loss, i.e.,
\[
  l(\btheta_1,\btheta_2,\btheta_3) \triangleq
  l_1(\btheta_1) + \alpha l_2(\btheta_2) + \beta l_3(\btheta_3),
\]
where $\alpha$ and $\beta$ are hyperparameters controlling whether the
optimization goal is biased towards prediction accuracy or earliness.
The pseudo-code of the training procedure is shown in \cref{alg:train}.

\begin{algorithm}
  \caption{Model training of KVEC\label{alg:train}}
  Initialize learning rates $\gamma_{\btheta}$ and $\gamma_{\btheta_b}$\;
  Randomly initialize model parameters $\btheta\triangleq(\btheta_1,\btheta_\pi)$ and $\btheta_b$\;
  \For{every tangled key-value sequence $\CS$ in the training dataset}{\label{ln:for_start}
      $n_k\gets 0$ for all $k \in \CS$\;
      \tcp{Generate episodes.}
      \For{each time step $t$ in the range from $1$ to $|\CS|$}{
        $e \gets$ item $\pr{k,\bv}$ at time step $t$\;
        $k\gets e.k$\;
        \lIf{sequence $\CS_k$ is already halted}{continue}
        $n_k\gets n_k +1$\;
        $\bs_k^{(t)}\gets$ update the representation of sequence $\CS_k$\;
        Sample an action $a_k^{(t)}\sim\pi(\bs_k^{(t)})$\;
        \If{$a_k^{(t)}=\mathrm{Halt}$}{
          $\hat{y}_k\gets$ classifier predicts the label of $\CS_k$ based on current $\bs_k^{(t)}$\;
          \uIf{$\hat{y}_k=y_k$}{$r_k^{(i)}\gets +1$ for $1\leq i\leq n_k$\; }
          \Else{$r_k^{(i)}\gets -1$ for $1\leq i\leq n_k$\;} \label{ln:for_end}
        }
      }
      \tcp{Update parameters based on episodes.}
       $\btheta \gets \btheta - \gamma_{\btheta} \nabla_{\btheta} l$\; \label{ln:update_start}
       $\btheta_b \gets\btheta_b - \gamma_{\btheta_b}\nabla_{\btheta_b}
       \mathrm{MSE}(b,R)$\; \label{ln:update_end}
    }
\end{algorithm}

Our training dataset consists of a number of tangled key-value sequences.
The training process consists of two stages.
First, we generate episodes for each tangled key-value sequence following a
halting policy $\pi$ (Lines~\ref{ln:for_start} to~\ref{ln:for_end}).
Then, we update parameters based on these episodes using different learning rates
(Lines~\ref{ln:update_start} to~\ref{ln:update_end}).
Note that $\btheta_b$ is updated independently via a regression aiming to minimize
the mean squared error (MSE) between accumulated reward $R$ of the episode and
state-value function.

\section{Experiments}
\label{sec:experiments}

In this section, we conduct experiments on four real-world key-value sequence
datasets and a synthetic dataset to evaluate the performance of our proposed KVEC framework.

\subsection{Experimental Setup}

\subsubsection{Datasets}

Our datasets include two common publicly available datasets, two self-collected
datasets, and a synthetic dataset, which are described below.

\header{USTC-TFC2016}~\cite{Wang2017} is a collection of network traffic traces
for malware and intrusion detection.
We discard short network flows with less than ten packets, and finally retain four
benign application types and five malicious application types.
We represent the dataset as a tangled key-value sequence as follows.
Each item in the sequence represents a packet.
The key of an item is the five-tuple of packets, and the value of an item is a
two-dimensional vector, i.e., packet size and packet direction (from client to
server or from server to client).
Items are then mixed chronologically to form a tangled key-value sequence.

\header{MovieLens-1M}~\cite{MovieLens1M} is a movie rating dataset containing one
million rating records, generated by $6,040$ users rating on $3,900$ movies.
Each user in the dataset has a profile, including his gender, age, and occupation.
We convert this data to a tangled key-value sequence as follows.
Each item in the sequence represents a user-movie rating record.
The key of an item is the user ID, and the value of an item is a three-dimensional
vector, i.e., movie ID, movie genre, and rating.
Each user's gender is the label that we want to predict.
Items are organized in chronological order to form a tangled key-value sequence.

\header{Traffic-FG} is a network traffic dataset collected from our campus
backbone network in 2020.
The dataset contains fine-grained encrypted traffic data with twelve categories of
service-level TCP network flows.
The processing of this dataset is same to the USTC-TFC2016 dataset.

\header{Traffic-App} is another self-collected network traffic dataset containing
application-level encrypted traffic data with ten application types, where six of
them are TCP applications and four of them are UDP applications.
The processing of this dataset is same to the USTC-TFC2016 dataset.

\header{Synthetic-Traffic} is a synthetic network traffic dataset generated in
ours controllable setting.
As precise halting positions are typically unlabeled in real-world datasets, we
constructed this dataset to assess the performance of the halting policy in KVEC.
It consists of both an early-stop subdataset and a late-stop subdataset.
The true stop signal is positioned at the start (or end) of the packet sequence in
the early-stop (or late-stop) subdataset.
This design is informed by the observation that the first few packets in a network
flow carry crucial information for identifying it~\cite{Rezaei2019}.
We randomly select two classes of concurrent network flows from the Traffic-App
dataset, intercepting the first ten packets of each flow as the stop signal and
combining them with empty packets to generate this synthetic traffic dataset.

In the data processing, we define a session in the traffic datasets as
a continuous packet subsequence that maintains the same transmission direction
within a network flow (which aligns with the established concept of {\em burst}
in traffic analysis).
In the MovieLens-1M dataset, we define a session as a subsequence of movies with
the same genre that a user continuously watched.
The detailed statistics of these five datasets are provided in \cref{tab:data}.

\begin{table}[htp]
  \caption{Detailed statistics of each dataset.\label{tab:data}}
  \begin{center}
    \begin{tabular}{@{\,}c@{ }|c|c|@{ }c@{ }|c@{\,}}
      \hline
      \hline
      dataset      & \#keys  & avg $|\CS_k|$ & avg session length & \#classes \\
      \hline
      USTC-TFC2016 & $3,200$  & $31.2$        & $8.3$              & $9$       \\
      MovieLens-1M & $6,040$ & $163.5$       & $1.7$             & $2$       \\
      Traffic-FG   & $60,000$ & $50.7$        & $2.4$              & $12$      \\
      Traffic-App  & $50,000$ & $57.5$        & $2.7$              & $10$      \\
      Synthetic-Traffic  & $10,000$ & $100.0$        & $2.1$              & $2$      \\
      \hline
    \end{tabular}
  \end{center}
\end{table}

\subsubsection{Baseline Methods}

Since key-value sequence data is a special time series data, we consider using the
state-of-the-art time series early classification methods as our baselines.
Besides, we construct baselines that use the Transformer~\cite{Vaswani2017a} to
learn a representation for each key-value sequence independently without
considering the item correlations between sequences.
We refer to such baselines as sequence representation networks (SRN), and will
also construct its several variants.

\begin{itemize}
\item \textbf{EARLIEST}~\cite{Hartvigsen2019} is the state-of-the-art time series
  early classification method.
  It uses the LSTM recurrent neural network to model multivariate time series
  data, and uses a reinforcement learning based halting policy network to decide
  whether to stop and classify current time series, or wait for more data.

\item \textbf{SRN-EARLIEST}.
  We replace the LSTM module in EARLIEST by Transformer~\cite{Vaswani2017a} as
  Transformer is reported to be better than LSTM in many recent studies.
  We denote this modified EARLIEST algorithm by SRN-EARLIEST.

\item \textbf{SRN-Fixed}.
  Inspired by work~\cite{ma2016}, the simplest halting policy for early
  classification is to stop at a fixed time point.
  We combine this simple halting policy with SRN, and denote this baseline by
  SRN-Fixed.
  Here, the fixed time point will be considered as a hyperparameter that controls
  prediction earliness.

\item \textbf{SRN-Confidence}.
  Inspired by work~\cite{parrish2013}, we propose a confidence based halting
  policy.
  In this method, the input sequence is halted when the classifier's confidence
  score for the prediction exceeds a predetermined confidence threshold $\mu$.
  We combine this halting policy with SRN, and denote this baseline by
  SRN-Confidence.
  Here, threshold $\mu$ will be considered as a hyperparameter that controls
  prediction earliness.
\end{itemize}

All above early classification baselines have some hyperparameters, which are
summarized in \cref{tab:hyper}, and their main function is balancing the
prediction accuracy and earliness.

\begin{table}[htp]
  \centering
  \caption{Hyperparamets of different early classification methods.\label{tab:hyper}}
  \begin{tabular}{c|c|c}
    \hline
    \hline
    method         & hyperparameters   & description                  \\
    \hline
    KVEC           & $\alpha,\beta$ & earliness-accuracy trade off \\
    EARLIEST       & $\lambda$         & earliness-accuracy trade off \\
    SRN-EARLIEST   & $\lambda$         & earliness-accuracy trade off \\
    SRN-Fixed      & $\tau \geq 1$     & halting time threshold       \\
    SRN-Confidence & $\mu \in [0,1]$   & halting confidence threshold \\
    \hline
  \end{tabular}
\end{table}

\subsubsection{Performance Metrics}


In order to conduct a comprehensive comparison study, we will also evaluate
different methods using the following metrics.

\header{Earliness}.
The earliness measures how quickly a method can correctly classify a sequence.
It is defined as follows:
\[
  \mathrm{Earliness}\triangleq\frac{1}{K}\sum_{k=1}^K\frac{n_k}{|\CS_k|}
\]
where $K$ is the total number of key-value sequences, and $n_k$ is the number of
observed items in key-value sequence $\CS_k$.

\header{Classification Performance Metrics}.
Classical classification performance metrics including accuracy, precision,
recall, and F1 score are also used, which are defined below.
\begin{align*}
  \mathrm{Accuracy}&\triangleq\frac{1}{K}\sum_{k=1}^K\indr{y_k=\hat{y}_k}, \\
  \mathrm{Precision} &\triangleq \frac{\text{TP}}{\text{TP} + \text{FP}}, \\
  \mathrm{Recall} &\triangleq \frac{\text{TP}}{\text{TP} + \text{FN}}, \\
  \text{F1 score} &\triangleq
  \frac{2\times(\mathrm{Precision}\times\mathrm{Recall})}{\mathrm{Precision}+\mathrm{Recall}}.
\end{align*}
Here, $y_k$ and $\hat{y}_k$ are the ground truth label and predicted label for
sequence $\CS_{k}$, respectively.
TP, FP, and FN denote the number of true positives, false positives, and false
negatives, respectively.

\header{Harmonic Mean of Accuracy and Earliness (HM)}.
Inspired by the definition of F1 score, we use the harmonic mean of accuracy and
earliness to measure the multi-objective balancing ability of different methods,
defined by
\[
  \mathrm{HM} \triangleq
  \frac{2\times(1-\mathrm{Earliness})\times\mathrm{Accuracy}}
  {1-\mathrm{Earliness}+\mathrm{Accuracy}}.
\]

Note that HM is a score between $0$ and $1$.
The larger HM is, the better an method performs.

\subsubsection{Settings}
We split each dataset into training, validation, and test subdatasets
with proportion $8:1:1$ based on the key field of items.
To prevent information leakage from test samples during training, there is no
intersection between keys in different subdatasets.
We conduct five-fold cross-validation on each dataset and report the average
performance.

In the KVRL module, $6$ (or $2$) stacked attention blocks are employed to learn
$128$-dimensional (or $64$-dimensional) item embeddings on the network traffic
(or MovieLens-1M) datasets.
Each attention block includes an Attention Mechanism layer and a Feed-Forward layer with ReLU 
nonlinear activation function and $0.1$ dropout probability.
Subsequently, a single LSTM layer with $256$ cells is utilized to learn the
sequence embedding in embedding fusion block.
During the training phase, the learning rate is set to $10^{-4}$ for the network
traffic datasets and $10^{-3}$ for the MovieLens-1M dataset.
Each method is run for $100$ epochs with a batch size of $64$, using the Adam
optimizer.
All datasets and model implementations are publicly available at
\href{https://github.com/tduan-xjtu/kvec\_project}{https://github.com/tduan-xjtu/kvec\_project}.

\subsection{Results}

It is worth noting that the early classification problem is a
multi-objective optimization task.
One method may have good accuracy but large delay, and the
other method may have poor accuracy but small delay.
To conduct a fair comparison, in experiments, we tune the hyperparameters of 
each method in \cref{tab:hyper} to
obtain the performance-earliness curve, and then compare the performance of 
different methods under the same earliness level.


\begin{figure}[htp]
  \centering
  \subfloat[USTC-TFC2016]{\includegraphics[width=.5\linewidth]{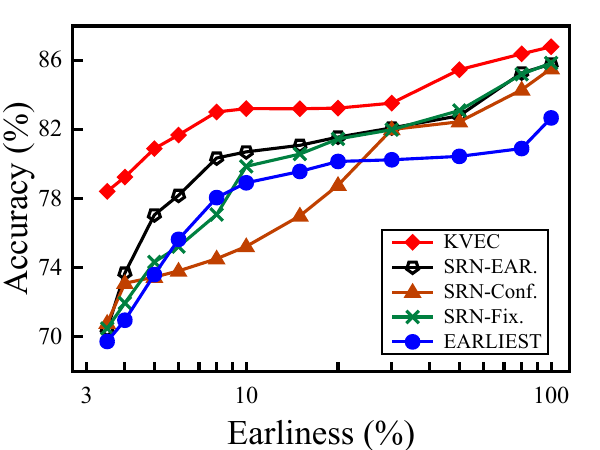}}
  \subfloat[MovieLens-1M]{\includegraphics[width=.5\linewidth]{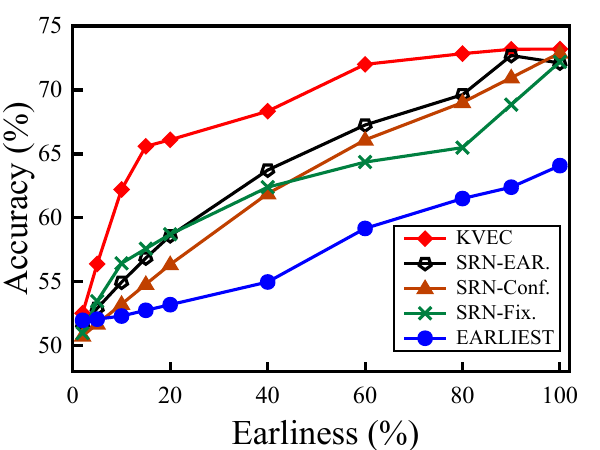}}\\
  \subfloat[Traffic-FG]{\includegraphics[width=.5\linewidth]{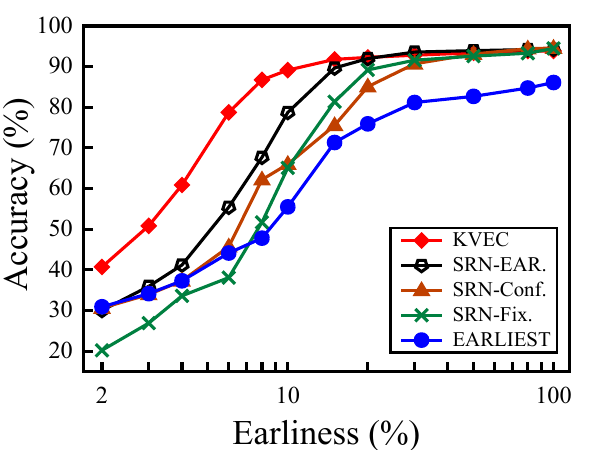}}
  \subfloat[Traffic-App]{\includegraphics[width=.5\linewidth]{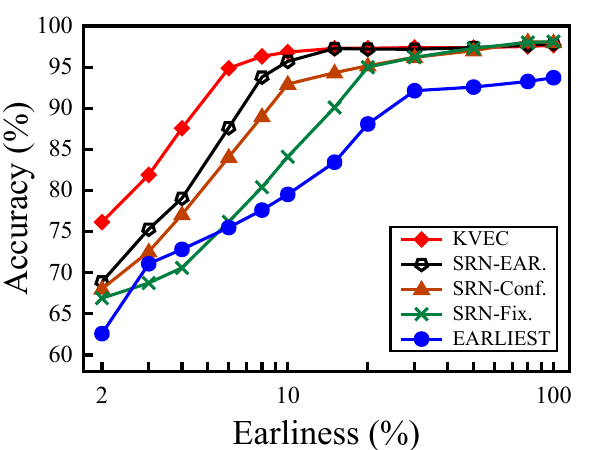}}
  \caption{Accuracy comparison\label{fig:accuracy}}
  \vspace{-2ex}
\end{figure}

\begin{figure}[htp]
  \centering
  \subfloat[USTC-TFC2016]{\includegraphics[width=.5\linewidth]{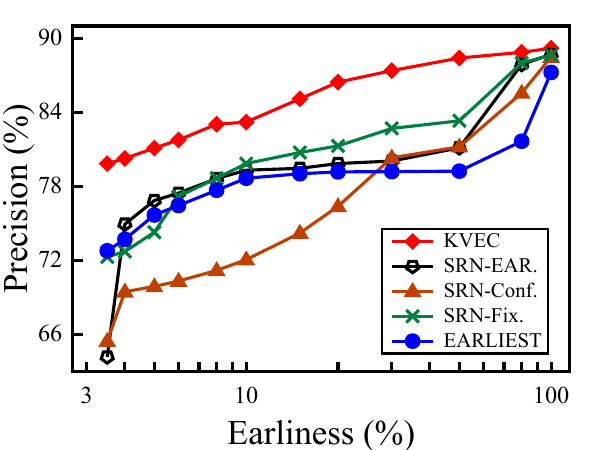}}
  \subfloat[MovieLens-1M]{\includegraphics[width=.5\linewidth]{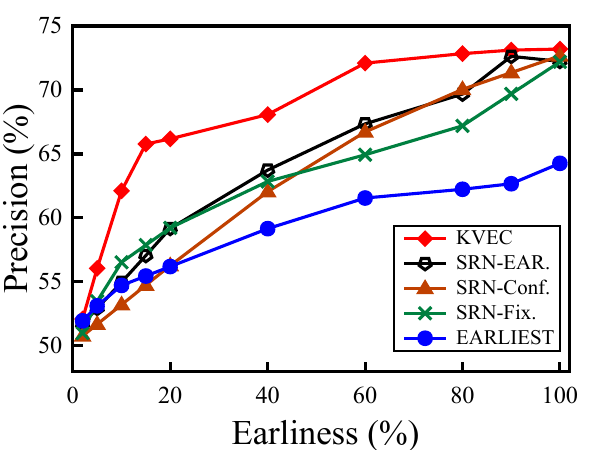}}\\
  \subfloat[Traffic-FG]{\includegraphics[width=.5\linewidth]{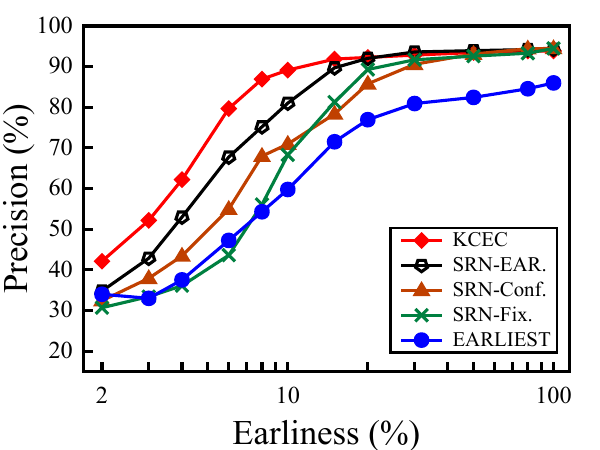}}
  \subfloat[Traffic-App]{\includegraphics[width=.5\linewidth]{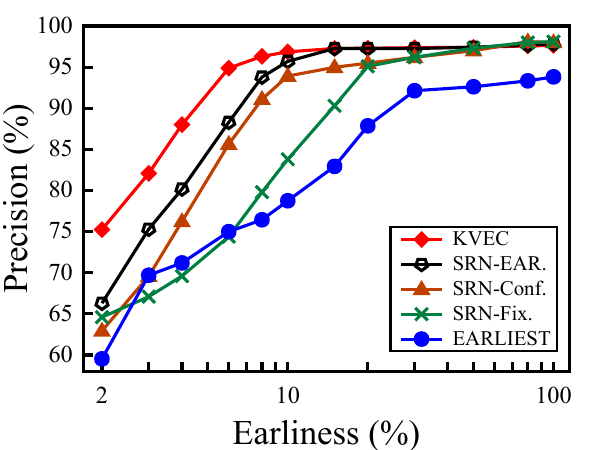}}
  \caption{Precision comparison\label{fig:precision}}
  \vspace{-2ex}
\end{figure}

\begin{figure}[htp]
  \centering
  \subfloat[USTC-TFC2016]{\includegraphics[width=.5\linewidth]{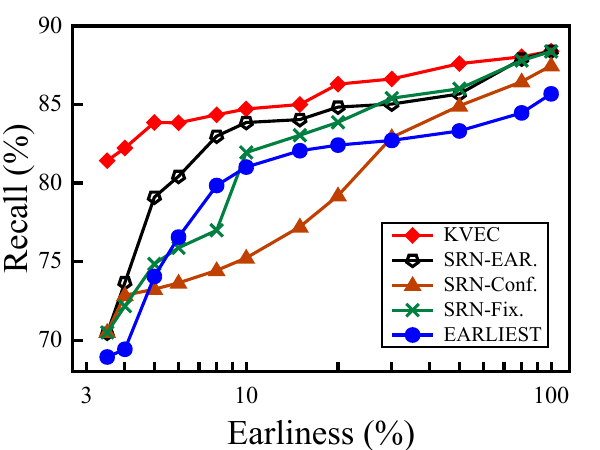}}
  \subfloat[MovieLens-1M]{\includegraphics[width=.5\linewidth]{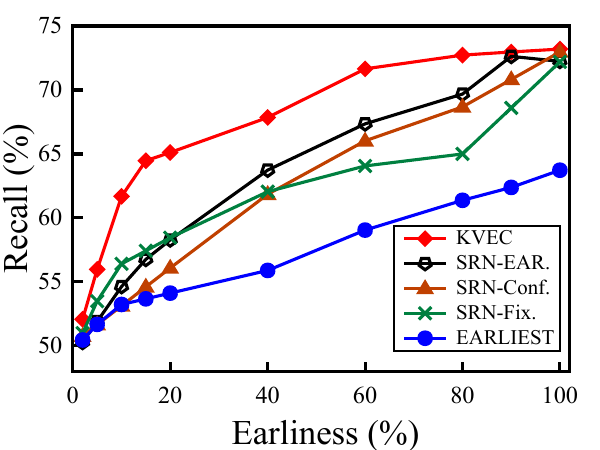}}\\
  \subfloat[Traffic-FG]{\includegraphics[width=.5\linewidth]{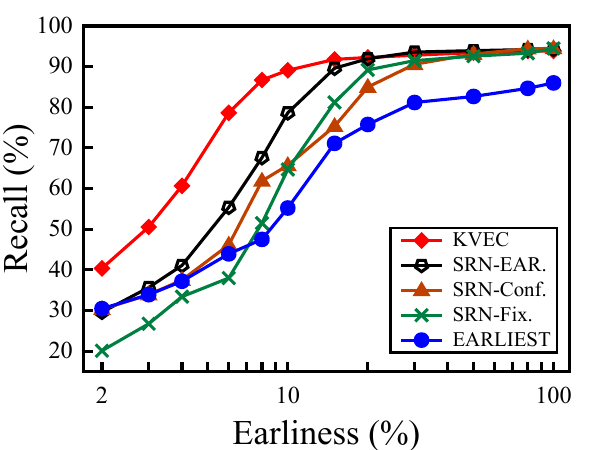}}
  \subfloat[Traffic-App]{\includegraphics[width=.5\linewidth]{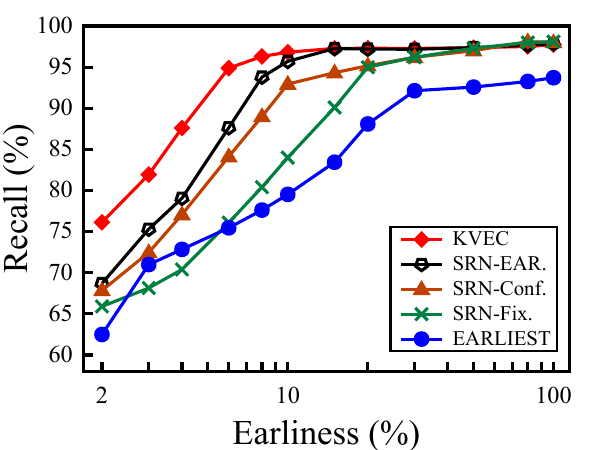}}
  \caption{Recall comparison\label{fig:recall}}
  \vspace{-2ex}
\end{figure}

\begin{figure}[htp]
  \centering
  \subfloat[USTC-TFC2016]{\includegraphics[width=.5\linewidth]{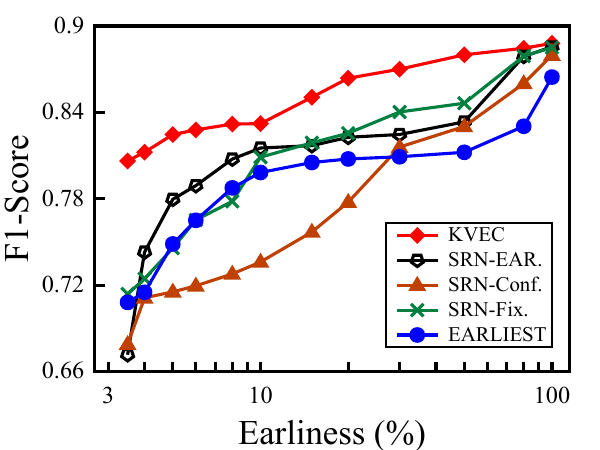}}
  \subfloat[MovieLens-1M]{\includegraphics[width=.5\linewidth]{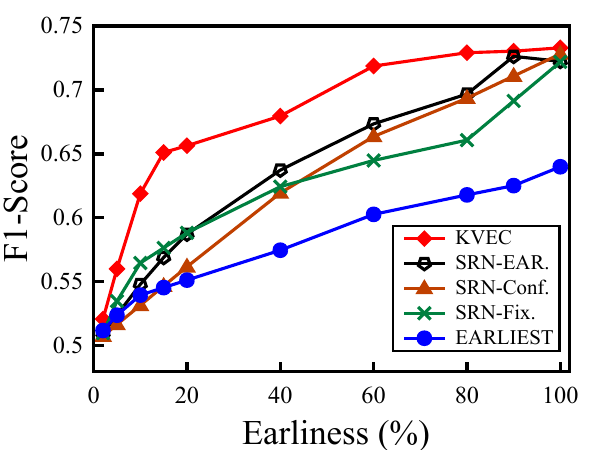}}\\
  \subfloat[Traffic-FG]{\includegraphics[width=.5\linewidth]{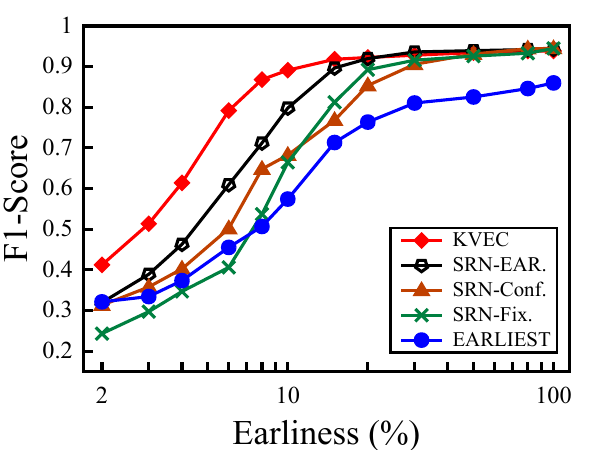}}
  \subfloat[Traffic-App]{\includegraphics[width=.5\linewidth]{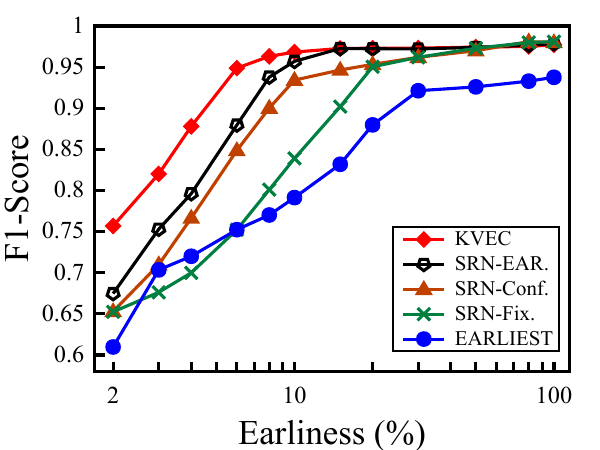}}
  \caption{F1-score comparison\label{fig:F1}}
  \vspace{-2ex}
\end{figure}

\begin{figure}[htp]
  \centering
  \subfloat[USTC-TFC2016]{\includegraphics[width=.5\linewidth]{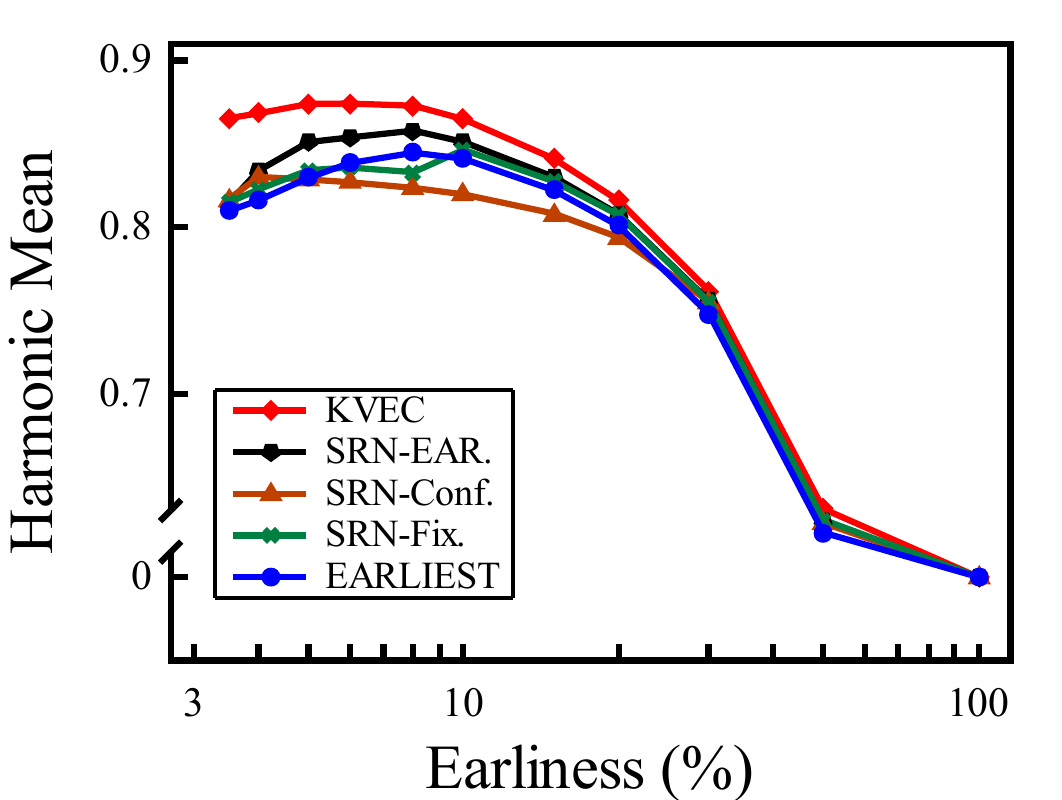}}
  \subfloat[MovieLens-1M]{\includegraphics[width=.5\linewidth]{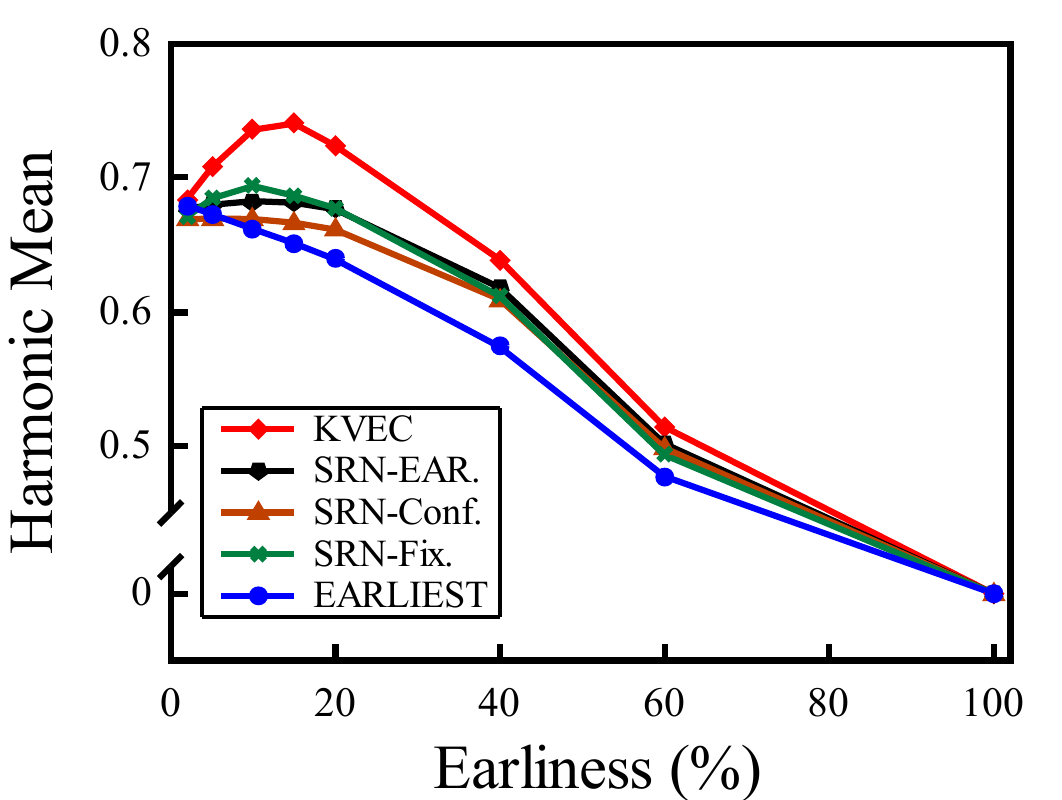}}\\
  \subfloat[Traffic-FG]{\includegraphics[width=.5\linewidth]{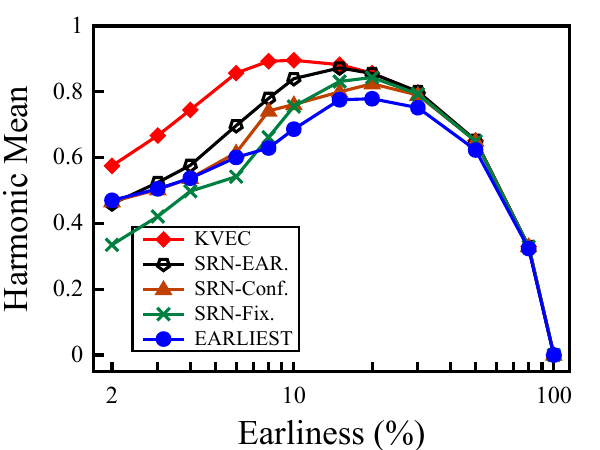}}
  \subfloat[Traffic-App]{\includegraphics[width=.5\linewidth]{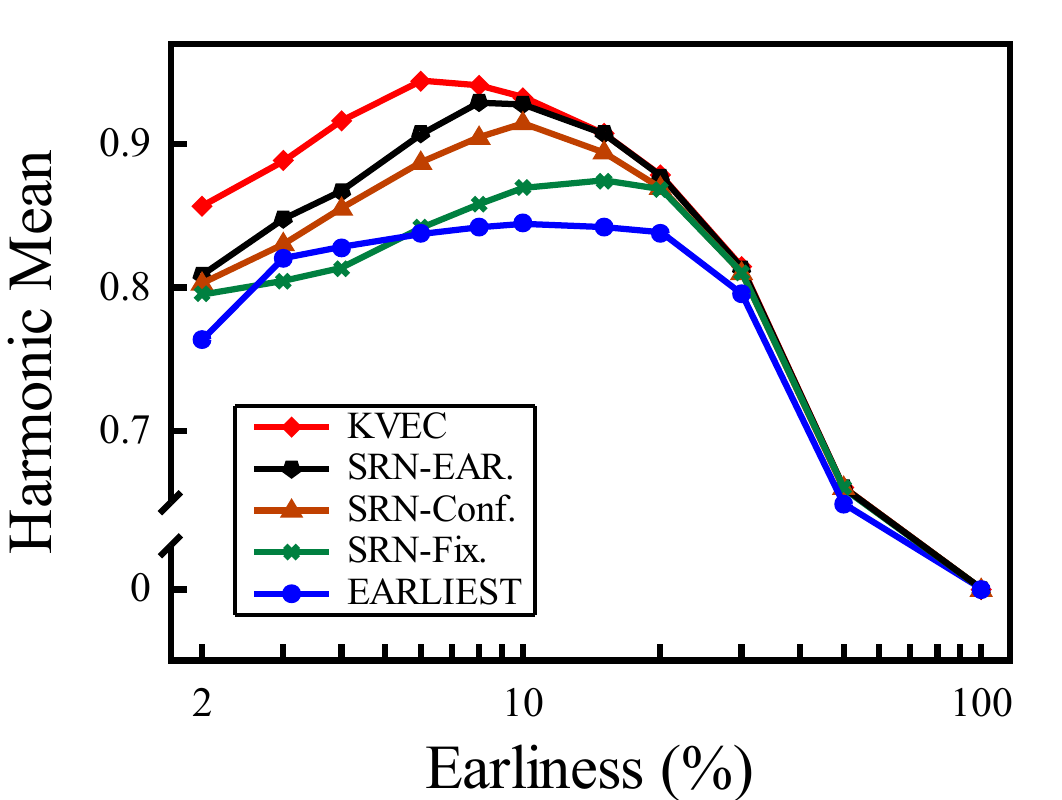}}
  \caption{HM comparison\label{fig:HM}}
  \vspace{-2ex}
\end{figure}

Figures~\ref{fig:accuracy} to \ref{fig:F1} depict the curves of classification
performance vs.~earliness of each early classification method on the four real-world
datasets. We observe that the metrics increase as the earliness
becomes large.
This observation is consistent with our intuition, as more items are observed in a
key-value sequence, we are more confident about its class label and hence achieve
better classification performance.
Comparing KVEC with other methods, we observe that, on all of real-world datasets,
KVEC consistently achieves the highest accuracy than the others.
This improvement is particularly notable in the early period, i.e., earliness is
small.
Specifically, when earliness is in the range of $2 - 8\%$, KVEC achieves
accuracy improvements of $4.7\%$, $17.5\%$, and $6.4\%$ on the three traffic
datasets, respectively, in comparison with the most competitive baseline
SRN-EARLIEST; and achieves an average accuracy improvement of $7.8\%$ on the
MovieLens-1M dataset than SRN-EARLIEST when the earliness is in the range of
$10 - 40\%$.
Similar performance improvements are observed for precision, recall, and F1 score.
This observation demonstrates that our KVEC has a better classification
performance than the other methods, especially when prediction earliness is
crucial.

Furthermore, we observed that EARLIEST, despite being considered the
state-of-the-art TSEC method, exhibits nearly the poorest performance on
key-value sequence datasets.
Even when the entire sequence is observed (i.e., earliness equals $100\%$), the
accuracy of EARLIEST remains significantly lower than other methods.
This disparity stems from the inherent differences between time series data and
key-value sequence data, affirming our assertion that existing TSEC methods are
not well-suited for key-value sequence data.

Figure~\ref{fig:HM} illustrates the relation between harmonic mean and earliness
of each early classification method.
On the four real-world datasets, we observe that the curves first increase and then
drop as earliness increases.
This is due to the property of harmonic mean in its definition.
Similar to the observations in \cref{fig:accuracy}, on the four real-world datasets,
KVEC generally performs better than the other methods, and the average
improvements on the four datasets are about $2.9\%, 5.3\%, 14.0\%$, and $3.7\%$,
respectively, with respect to the best among other baselines.
This indicates that KVEC achieves a better balance between earliness and accuracy,
compared to other baselines.

\subsection{Hyperparameter Sensitivity}

\begin{figure}[htp]
  \centering
  \subfloat[Effect of $\alpha$ on performance\label{Fig.alpha}]{%
    \includegraphics[width=0.5\linewidth]{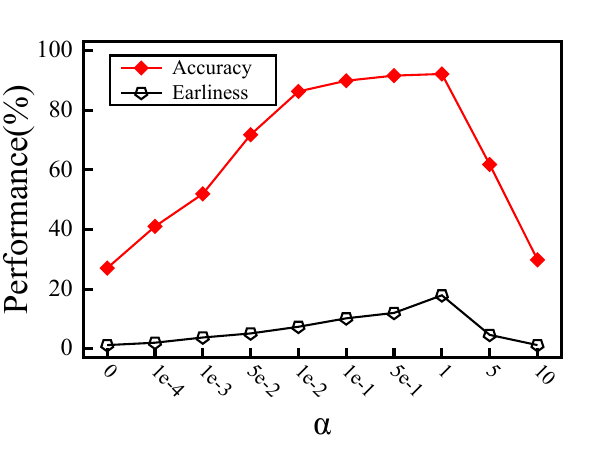}}
  \subfloat[Effect of $\beta$ on performance\label{Fig.beta}]{%
    \includegraphics[width=0.5\linewidth]{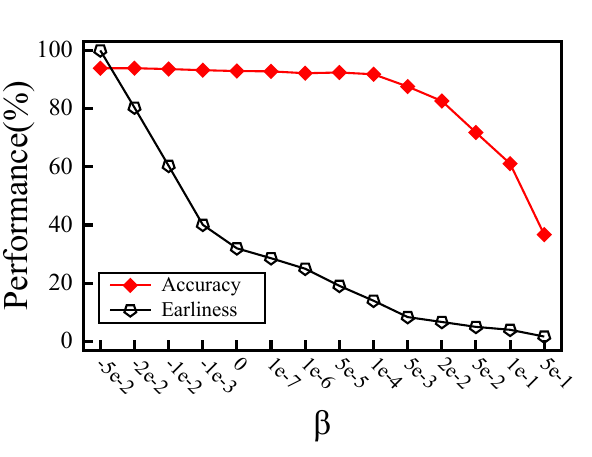}}\\
  \caption{Hyperparameter sensitivity analysis.}
  \label{fig:hyperparameter}
\end{figure}

Next, we study the sensitivity of the hyperparameters in KVEC, i.e., $\alpha$ and
$\beta$, which regulate the weight of halting policy and time penalty to the
overall loss function, respectively.
We illustrate the hyperparameters' effects on prediction accuracy and earliness on
the Traffic-FG dataset in \cref{fig:hyperparameter}.

In \cref{Fig.alpha}, we maintain $\beta$ at $10^{-4}$ and tune $\alpha$
within the range of $[0,10]$ to control the update weight of the halting policy.
Instead, in \cref{Fig.beta}, we set $\alpha$ to $0.1$ and tune $\beta$ within
the range of $[-0.05, 5]$ to control the intensity of the time delay penalty.
We observe that $\alpha$ significantly impacts accuracy but has little effect on
earliness; while $\beta$ primarily serves to balance prediction accuracy and
earliness by penalizing halting time delay.
According to this observation, in the experiments, we freeze $\alpha$ at $0.1$
and tune $\beta$ within the range of $[-0.05, 5]$ to obtain the complete
performance curve of KVEC.

\subsection{Ablation Study}

\begin{figure}[htp]
  \centering
  \subfloat[Effect on Accuracy]{\includegraphics[width=0.5\linewidth]{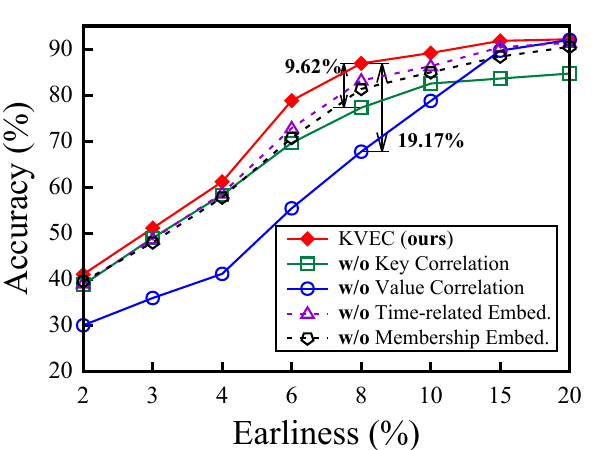}}
  \subfloat[Effect on Harmonic Mean]{\includegraphics[width=0.5\linewidth]{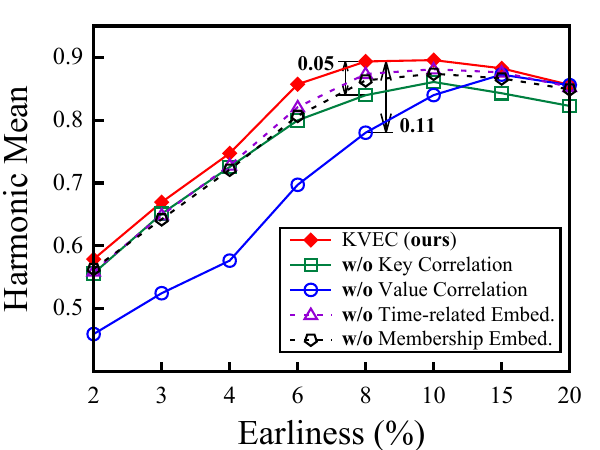}}
  \caption{Ablation study\label{fig:ablation}}
\end{figure}

We perform an ablation study to investigate the function of each component in KVEC.
The experiments are conducted on the Traffic-FG dataset, and the results are shown
in \cref{fig:ablation}.

In the figure, ``w/o Key Correlation'' means that we remove the key correlation,
only the value correlation is preserved in KVEC; ``w/o Value Correlation'' means
that we completely remove the value correlation in KVEC, and treat each key-value
sequence independently; ``w/o Time-related Embed.'' means that all time-related
embeddings including relative position embedding and time embedding in the input
embedding are completely removed; ``w/o Membership Embed.'' means that the
membership embedding in input embedding is removed.

We have the following observations.
First, the performance of the KVEC has the most significant slump after removing
the value correlation.
This confirms that the powerful performance of KVEC is mainly due to value
correlations in a tangled key-value sequence.
Meanwhile, removing the key correlation also harms the performance of KVEC.
Second, the performance of KVEC without relative position embedding and time
embedding performances drops slightly, which indicates that time-related
information indeed can improve the quality of sequence representation learning and
timing learning.
Third, the performance of KVEC without the membership embedding also drops
slightly, which indicates that distinguishing correlation information from
different sequences is essential for learning a better sequence representation.

\subsection{Discussions}

  In this section, to better understand the proposed KVEC framework, we conduct a
  comprehensive qualitative analysis to answer the following research questions:
  \begin{itemize}
  \item \textbf{RQ1}: How does the KVRL module actually work, and does its
    attention mechanism indeed benefit early classification?
  \item \textbf{RQ2}: How does the halting policy work, and does the KVRL
    module effectively facilitates early halting?
  \item \textbf{RQ3}: What are the reliability and limitations of KVEC in practical application scenarios?
  \end{itemize}

\begin{figure}[htp] \centering
    \includegraphics[width=0.75\linewidth]{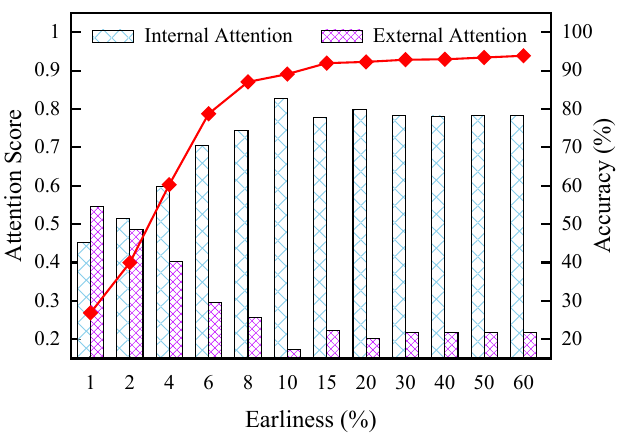}
    \caption{Distribution of attention score in various halting positions.}
    \label{fig:attention}
  \end{figure}

  \header{Attention mechanism in KVEC (RQ1).}
  In the KVRL module, a pivotal design is a novel attention mechanism that
  leverages item correlations from both intra-sequence and inter-sequence.
  To answer RQ1, we quantify the attention distribution of KVRL module
  under various halting positions.
  In this experiment, we specifically employ the internal attention score and
  external attention score to represent the cumulative attention weights derived
  from key correlations and value correlations, respectively.
  We conduct this experiment using the Traffic-FG dataset and compare the attention
  scores and accuracy at various earliness, as depicted in \cref{fig:attention}.

  We observe that, the external attention score in the early time period (i.e.,
  earliness is less than $10\%$) is higher than in the later time period.
  When more items in the sequence are observed, the external attention score
  generally decreases, and internal attention score progressively assumes
  dominance.
  This indicates that in early classification tasks, 
  since intra-sequence data is insufficient, KVEC prioritizes exploring
  inter-sequence correlations to enhance sequence representations, thereby improving the earliness.
  In contrast, when sufficient data is observed, KVEC tends to focus on exploiting
  intra-sequence correlations to achieve higher accuracy.

\begin{figure}[htp]
  \centering
  \subfloat[Early-stop subdataset\label{Fig.L_halt}]{%
    \includegraphics[width=0.5\linewidth]{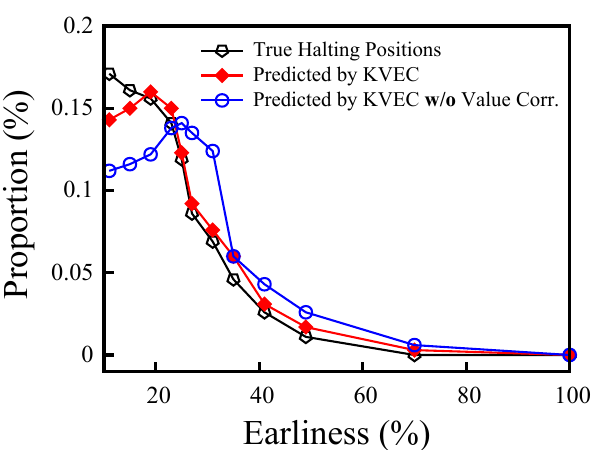}}
  \subfloat[Late-stop subdataset\label{Fig.R_halt}]{%
    \includegraphics[width=0.5\linewidth]{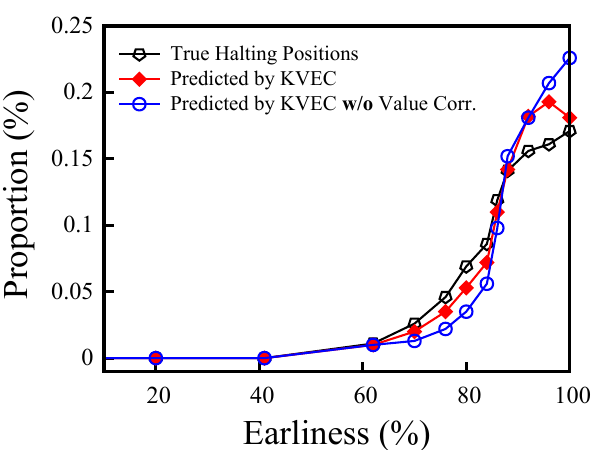}}
  \caption{Distribution of halting positions predicted by different methods.}
  \label{fig:halt}
\end{figure}

  \header{Halting policy in KVEC (RQ2).}
  In KVEC, a reinforcement learning-based halting policy is employed to control
  the timing of early classification.
  To answer RQ2, we compare the disparity between the halting
  positions predicted by the halting policy in KVEC and its variant (i.e., KVEC
  without value correlation) and the true halting positions
  within the Synthetic-Traffic dataset.
  The results are depicted in \cref{fig:halt}.

We observe that the distribution of halting positions 
in both early-stop data and late-stop data
predicted by KVEC are closest to 
the distribution of true halting positions.
This indicates that the halting policy within KVEC can accurately capture dynamic
stopping signals, and proposed representation learning method (i.e., KVRL) indeed
facilitates this ability.

\begin{figure}[htp]
  \centering
  \subfloat[Accuracy vs.~earliness]{%
    \includegraphics[width=0.5\linewidth]{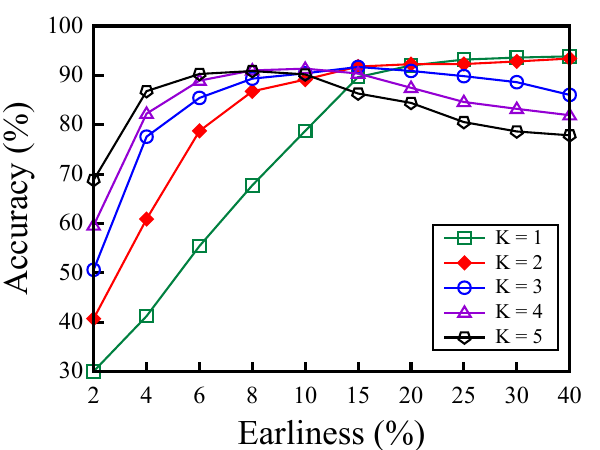}}
  \subfloat[Harmonic mean vs.~earliness]{%
    \includegraphics[width=0.5\linewidth]{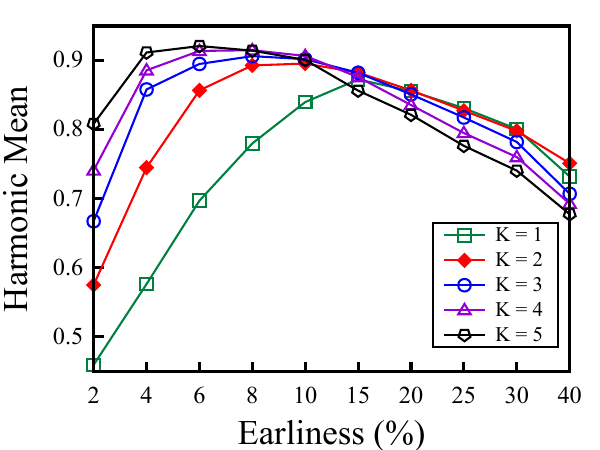}}
  \caption{Effects of $K$ to performance of KVEC.}
  \label{fig:N}
\end{figure}

\header{Performance in practical applications (RQ3).}
In KVEC, multiple concurrent key-value sequences within a tangled key-value sequence
are jointly modeled. In practical applications, the number of concurrent sequences in
various scenarios is a primary factor affecting KVEC's performance.
To answer question RQ3, we conduct experiments on the Traffic-FG dataset,
selecting diverse testing scenarios with varying numbers of concurrent key-value
sequences $K$. The results are depicted in \cref{fig:N}.

Our observations reveal that, in various testing scenarios, KVEC exhibits higher
early classification accuracy at the early time period (i.e., earliness is less
than $10\%$) when $K$ becomes larger. However, while at the later time period,
KVEC performs worse when $K$ becomes larger.
This demonstrates that in a more complex scenario, i.e., with a larger
$K$, KVEC can exploit more inter-sequence correlations to enrich the sequence
representation, but these inter-sequence
correlations also inevitably introduce more noise information and lead to unstable
performance.
This phenomenon inspires us to find a more intelligent approach to leverage
inter-sequence correlations in future work.

\section{Conclusion}
\label{sec:conclusion}

In this work, we formulate a novel tangled key-value sequence early classification
problem, which has a wide range of applications in real world.
We propose the KVEC framework to solve this problem. 
KVEC mainly consists of two steps, i.e., the key-value sequence representation 
learning, and the joint optimization of prediction accuracy and prediction earliness.
The core idea behind KVEC is to leverage the rich intra- and inter-sequence 
correlations within the tangled key-value sequence to obtain an informative sequence 
representation.
Through combining representation learning with timing learning, KVEC dynamically 
determines the optimal timing for early classification.
Extensive experiments conducted on both real-world and synthetic datasets 
demonstrate that KVEC outperforms all alternative methods.

\section*{Acknowledgements}
\label{sec:acknowledgements}

We are grateful to anonymous reviewers for their constructive comments
to improve this paper. 
This work was supported in part by National Natural Science 
Foundation of China (62272372, 61902305, U22B2019).

\clearpage
\bibliographystyle{IEEEtran}
\bibliography{IEEEabrv,ref}


\end{document}